\documentclass[10pt,twocolumn,letterpaper]{article}

\usepackage[pagenumbers]{cvpr} 

\usepackage[dvipsnames]{xcolor}
\usepackage{lipsum}

\definecolor{cvprblue}{rgb}{0.21,0.49,0.74}
\usepackage[pagebackref,breaklinks,colorlinks,citecolor=cvprblue]{hyperref}

\usepackage[utf8]{inputenc} 
\usepackage[T1]{fontenc}    
\usepackage{hyperref}       
\usepackage{url}            
\usepackage{booktabs}       
\usepackage{amsfonts}       
\usepackage{nicefrac}       
\usepackage{microtype}      
\usepackage{multirow}
\usepackage{graphicx}
\usepackage{subcaption}
\usepackage{amsmath}
\usepackage{xcolor}
\usepackage{tabularx}

\usepackage{colortbl}

\def\paperID{16} 
\def\confName{CVPR-W}
\def\confYear{2024}

\title{Uncovering the Hidden Cost of Model Compression}

\author{Diganta Misra \thanks{equal contribution}\\
Carnegie Mellon University, Landskape AI\\
{\tt\small digantam@andrew.cmu.edu}
\and
Muawiz Chaudhary \footnotemark[1]\\
Mila - Quebec AI Institute, Concordia University
\and
Agam Goyal \footnotemark[1]\\
University of Wisconsin-Madison
\and
Bharat Runwal \footnotemark[1]\\
Mila - Quebec AI Institute
\and
Pin Yu Chen\\
IBM Research
}

\begin{document}

\maketitle

\begin{abstract}
In an age dominated by resource-intensive foundation models, the ability to efficiently adapt to downstream tasks is crucial. Visual Prompting (VP), drawing inspiration from the prompting techniques employed in Large Language Models (LLMs), has emerged as a pivotal method for transfer learning in the realm of computer vision. As the importance of efficiency continues to rise, research into model compression has become indispensable in alleviating the computational burdens associated with training and deploying over-parameterized neural networks. A primary objective in model compression is to develop sparse and/or quantized models capable of matching or even surpassing the performance of their over-parameterized, full-precision counterparts. Although previous studies have explored the effects of model compression on transfer learning, its impact on visual prompting-based transfer remains unclear. This study aims to bridge this gap, shedding light on the fact that \textbf{model compression detrimentally impacts the performance of visual prompting-based transfer}, particularly evident in scenarios with low data volume. Furthermore, our findings underscore the adverse influence of sparsity on the calibration of downstream visual-prompted models. However, intriguingly, we also illustrate that such negative effects on calibration are not present when models are compressed via quantization. This empirical investigation underscores the need for a nuanced understanding beyond mere accuracy in sparse and quantized settings, thereby paving the way for further exploration in Visual Prompting techniques tailored for sparse and quantized models.
\end{abstract}    
\section{Introduction}
\label{sec:intro}

\begin{figure}[t]
  \centering
  \includegraphics[width=0.99\linewidth]{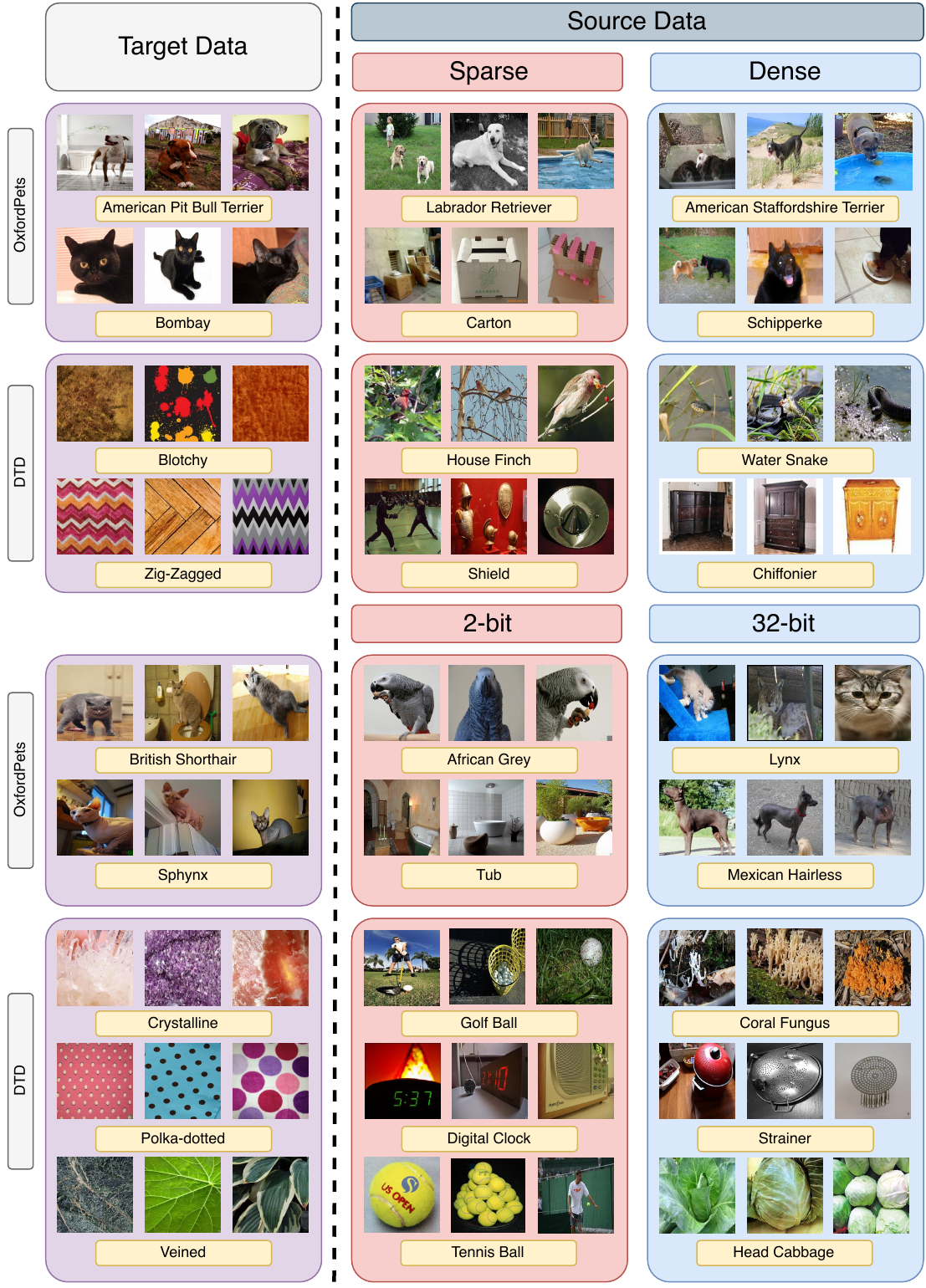}
  \caption{When examining the label mapping \cite{chen2023understanding}  of the ResNet-50 Sparse LT model \cite{he2016deep, frankle2018lottery} alongside its dense counterpart across target classes within the \textup{OxfordPets} \cite{parkhi12a} and \textup{DTD} \cite{cimpoi14describing} datasets, a notable distinction emerges: the dense model exhibits a more semantically accurate label mapping. In contrast, the sparse variant often assigns target classes to unrelated classes from the source dataset. This trend echoes similarly in the context of quantization, where the full-precision DeiT (32-bit) \cite{touvron2020training} demonstrates superior semantic accuracy and consistency in label mapping compared to its quantized counterpart (2-bit) \cite{huang2023variation} across various target classes within the \textup{OxfordPets} and \textup{DTD} datasets.}
  \label{fig:map}
  \vspace{-4mm}
\end{figure}

The evolution of deep learning has shifted from training task-specific models to extensive task-agnostic pre-training. The objective is to construct a model with robust universal representations, facilitating numerous downstream tasks without necessitating intensive training. This category of models is now encompassed by the term ``Foundation models'' (FM) \cite{bommasani2021opportunities}. The efficacy of these pre-trained models for downstream tasks has often been exemplified through straightforward and computationally efficient adaptation methods, such as linear probing (LP) \cite{alain2016understanding}, within the broader framework of transfer learning. Transfer learning has traditionally been fundamental for adapting pre-trained models to various downstream tasks, historically constrained to full fine-tuning (FF) and LP. The former, while generally superior in performance, is also the more costly approach, whereas the latter is a more computationally cheaper option but typically exhibits lower performance compared to full fine-tuning methods.

Although linear probing (LP) and full fine-tuning (FF) have traditionally served as standard transfer learning methods, a novel approach, referred to as model reprogramming \cite{elsayed2018adversarial, chen2022model, salman2021unadversarial} or more commonly known as visual prompting (VP) \cite{bahng2022exploring, tsai2020transfer, zhang2022fairness}, has emerged as a viable and efficient alternative, capable of rivaling LP in both performance and transfer cost. Essentially, VP involves learning a perturbation such that, when applied to input samples of the target dataset, the pre-trained model accurately classifies the resulting ``reprogrammed'' sample without requiring any changes to the weights. Reprogramming relies on aligning the features of the target data with those of the source data, eliminating the need for gradient updates to the network weights. Consequently, VP often competes with LP in terms of efficiency.


While several studies in the literature have empirically evaluated the performance of Linear Probing (LP) and Visual Prompting (VP) across diverse target data scenarios \cite{chen2022model, bahng2022exploring, yang2021voice2series, vinod2023reprogramming, tsao2023autovp}, there has been limited exploration of the distinctions between the two methods when subjected to the constraints of (a) \textit{low data volume} and (b) \textit{model compression}. This study aims to reveal the \textit{hidden costs} associated with compressed models in the domain of transfer learning across various data volume settings. In the existing literature, these costs span a broad spectrum, including model attributes such as reliability \cite{goldstein2020reliability, bernhard2019impact}, performance under distribution shifts\cite{chen2022sparsity, li2023learning}, fairness \cite{hooker2019compressed, hooker2020characterising, tran2022pruning}, and more.

To this end, under the constraint of low data volume, we examine the transferability of pre-trained models in few-shot settings. For the constraint of model compression, our investigation encompasses a wide array of sparse and quantized models generated through compression techniques such as unstructured pruning \cite{han2015deep,lecun1990optimal,han2015learning}, structural pruning \cite{liu2017learning,Zhou2016LessIM}, quantization~\cite{dettmers2023qlora,dettmers2022llmint8,frantar2023gptq,yao2022zeroquant,huang2023variation}, and solutions based on the Lottery Ticket Hypothesis (LTH) \cite{frankle2018lottery}.

Moreover, we expand our empirical framework to investigate the influence of transferring compressed models using VP on the calibration of the resulting model in the target task, comparing it to its dense counterparts. Although recent research has prioritized the comprehension and enhancement of fairness and reliability in models \cite{zhang2022fairness}, along with the examination of pruning strategies and the resultant models \cite{li2022calibration, gilles2020lottery, arora2023quantifying, lei2023calibrating, venkatesh2020calibrate}, there remains a gap in our understanding of the reliability of visual prompting as an adaptation mechanism, particularly in the context of varying model compression rates.

Our main contributions can be summarized as follows.

\begin{enumerate}
    \item We conduct a comprehensive empirical investigation into the effects of model compression and low data volume on the transferability of pre-trained models through visual prompting methods.
    \item Our in-depth analysis across eight datasets reveals a notable decrease in performance for compressed models compared to their dense counterparts when transferred via visual prompting.
    \item Significantly, we empirically examine the adverse impact of employing sparse models on the calibration of the final model obtained after transfer to the downstream target task using visual prompting, marking the first exploration of this phenomenon.
    \item Moreover, in contrast, we illustrate that quantized models do not suffer from the adverse effects on calibration observed in sparse models. We offer an intuitive analysis to elucidate this phenomenon.
    \item To that extent, we provide a fine-grained overview of our observations based on the categorization of different compression methods in Table \ref{tab:summary_res}. 
    \item Furthermore, we contribute novel insights into the distinctions between visual prompts and their corresponding label mapping learned by compressed models compared to their dense, full-precision counterparts. 
\end{enumerate}

\section{Background and Related Works}\label{sec:background}

\textbf{Visual Prompting and Reprogramming.}
Model reprogramming \cite{chen2022model,elsayed2018adversarial} is a novel parameter-efficient fine-tuning method that integrates two trainable modules, the input transformation layer and the output mapping layer, into a pre-trained model for transfer learning. During the reprogramming training process, only the parameters linked to the inserted layers are updated, maintaining the unchanged parameters of the pre-trained model. A significant advantage of model reprogramming lies in its proficiency in cross-domain learning, enabling successful application in diverse domains such as biomedical image classification \cite{tsai2020transfer}, time series classification in speech models \cite{yang2021voice2series}, and protein sequence learning in language models \cite{vinod2023reprogramming}. Specifically, in image classification tasks within the same domain, model reprogramming equates to visual prompting (VP) \cite{bahng2022exploring}. In simple terms, VP achieves input transformation through a universal trainable additive padding operation for each image and output mapping via a function specifying the transition from source label classes to target label classes. To realize the output mapping, \cite{tsai2020transfer} proposed frequency-guided label mapping, while \cite{chen2023understanding} proposed iterative label mapping for VP, known as ILM-VP. More description about VP are provided in supplementary material.

\textbf{Model Compression and Transfer Learning.} Model compression enhances inference efficiency by reducing memory requirements, with pruning and quantization standing out as prominent methods.

\textit{Pruning:} Achieving model sparsity, as introduced by \cite{lecun1990optimal,han2015deep}, effectively compresses over-parameterized deep neural networks. Progressive sparsification methods like GMP (Gradual Magnitude Pruning) \cite{DBLP:journals/corr/abs-1902-09574,han2015learning} and sparse regularization techniques such as AC/DC \cite{peste2021ac} and RigL \cite{evci2020rigging} showcase dynamic approaches to weight pruning. The Lottery Ticket Hypothesis (LTH) \cite{frankle2018lottery} identifies sparse subnetworks within large networks, offering comparable or superior performance. Recent work, like Upop \cite{shi2023upop}, addresses pruning limitations for large vision-language models. \textit{Quantization:} A widely used compression technique, quantization \cite{Gholami2021ASO}, involves representing weights or activations in lower precision. It includes quantization-aware training, conducted during fine-tuning or retraining, and post-training quantization applied after model training. Recent studies exploring quantization in LLM aim to reduce operation costs \cite{yao2022zeroquant,frantar2023gptq,dettmers2022llmint8}, albeit often at the expense of performance. For Vision Transformers, recent work \cite{huang2023variation} identifies unique variation behaviors in ViT, distinct from CNNs, and proposes an efficient knowledge distillation-based variation-aware quantization method to address this issue.
    
In transfer learning with sparse models, \cite{iofinova2022well} investigated various pruning techniques' effects on downstream performance using sparse pre-trained ImageNet models, comparing linear probing and fine-tuning for transfer. However, there is limited detailed study for quantized models in this context. Additionally, \cite{fu2023robust} introduced adversarial robustness into the transfer learning pipeline for improved transferability, while \cite{gorsline2021adversarial} explored the robustness of quantized models.

\textbf{Calibration of Neural Networks.} \cite{guo2017calibration} first identified neural networks' tendency for overconfident predictions, a phenomenon exacerbated in sparse training, such as \textup{RigL} \cite{evci2020rigging} as noted by \cite{lei2023calibrating}. While efforts focus on efficiency and adaptation speed in pruned models \cite{iofinova2022well}, sparse model reliability in transfer learning remains unexplored. Given their broad application and transfer across domains \cite{zhou2022domain}, overconfidence poses safety risks in critical domains like self-driving and healthcare \cite{hendrycks2021unsolved, hendrycks2022x}. Recent studies address calibration in pruned models \cite{arora2023quantifying}, propose sparse training improvements for lottery tickets \cite{lei2023calibrating}, and integrate calibration into pruning techniques \cite{venkatesh2020calibrate}.

Although visual prompt-based transfer has demonstrated promising outcomes across various downstream tasks, the dependability of this innovative method when employed with sparse models remains largely unexplored.
\section{Results}\label{sec:results}

\begin{table}[hbt!]
    \small
        \resizebox{1.0\linewidth}{!}{
    \begin{tabular}{c|c|c|c}
        \toprule
        Compression & Type & Models & Performance Drop \\
        \midrule
        GMP \cite{hagiwara1994simple, liu2021sparse} &  Unstructured &ResNet-(18,34) & High  \\
        IMP (LT) \cite{frankle2018lottery} &  Unstructured &  ResNet-50  & Moderate \\
        AC/DC \cite{peste2021ac} &  Unstructured& ResNet-50  & Low\\
        RigL \cite{evci2020rigging} &  Unstructured  & ResNet-50  & Low \\
        UPop \cite{shi2023upop} & Structured &CLIP ViT-L&High \\
        VVTQ \cite{huang2023variation} & Quantization & DeiT-T, Swin-T&Moderate \\
        \bottomrule
    \end{tabular}%
    }
    \caption{\textbf{Summary of Results.} Key Observations about the performance drop in various architectures of models using different compression strategies and transferred via Visual Prompting.}
    \label{tab:summary_res}
\end{table}

In this section, we look at results from our extensive analysis with the aim of understanding the difference between the downstream performance of various visual prompting methods in terms of performance under conditions of both low data volumes and model compression, as well as reliability in terms of calibration under different model compression rates. 

\subsection{Experimental Setup}

We consider eight target datasets that encompass a mixture of downstream tasks in the near- and far domain, namely \textup{CIFAR-10} \cite{krizhevsky2009learning}, \textup{SVHN} \cite{netzer2011reading}, \textup{GTSRB} \cite{Houben-IJCNN-2013}, \textup{DTD} \cite{cimpoi14describing}, \textup{Flowers102} \cite{Nilsback08}, \textup{OxfordPets} \cite{parkhi12a}, \textup{EuroSAT} \cite{helber2018introducing} and \textup{Caltech101} \cite{li_andreeto_ranzato_perona_2022}. In terms of model architectures, we base our experiments on the ResNet \cite{he2016deep}  family of models with ResNet-18, ResNet-34, ResNet-50 (\cref{sec:sparse_models}), CLIP \cite{radford2021learning} (\cref{sec:sparse_foundation_models}), and Vision Transformer \cite{DBLP:conf/iclr/DosovitskiyB0WZ21} models with DeiT-T and Swin-T \cite{touvron2020training, liu2021Swin} (\cref{sec:quantized_models}).


For pruned models, we use a GMP-pruned model \cite{hagiwara1994simple, liu2021sparse} for ResNet-18 and ResNet-34, and AC/DC \cite{peste2021ac}, RigL \cite{evci2020rigging}, and solutions of the lottery ticket hypothesis (LTH)\footnote{Solutions of LTH are essentially LT initializations that are trained to convergence.} \cite{frankle2018lottery} for ResNet-50 derived at different sparsity levels. All of these checkpoints are pre-trained on the ImageNet-1k \cite{deng2009imagenet} classification task. As shown in Fig. \ref{fig:LT}, most lottery ticket solutions demonstrate superior performance in terms of accuracy than that of the parent network on the pre-training dataset.

For quantized models, we use a VVTQuantized model \cite{huang2023variation} for DeiT-T and Swin-T models at varying bits of quantization on both activations and weights, from full 32 bit to 2 bit. All of these checkpoints are pre-trained on the ImageNet-1k \cite{deng2009imagenet} classification task. As mentioned in \cite{huang2023variation}, quantized 4 bit DeiT-T and Swin-T models obtain better performance than it's full-precision counterpart.

To ensure consistency and measure statistical significance, all configurations were run with three seeds, amounting to more than 15,000 experiments in total. Finally, for method of visual prompting (VP), we will be demonstrating our results for VP based on three popular label mapping techniques of Random Label Mapping (RLM-VP), Fequency-based Label Mapping (FLM-VP) \cite{tsai2020transfer}, and mainly Iterative Label Mapping (ILM-VP) \cite{chen2023understanding} which is the state-of-the-art VP method. Furthermore, for our study on the performance of sparse foundation models (\cref{sec:sparse_foundation_models}), we also compared ILM-VP with the implementation of visual prompting in \cite{bahng2022exploring}. Finally, we investigate the cross-modal reprogramming \cite{neekhara2022crossmodal} for various models and compression settings in \cref{sec:Cross Modal}.

\textbf{Note:} We define some terms and notations that will be used throughout this section:
\begin{itemize}
    \item $\mathbf{\Delta}$ in the LTH heat-maps (\cref{sec:lottery_ticket_hypothesis}) refers to the difference between the accuracy of the dense model and the corresponding LT at the specified sparsity and data budget. This same notation is also used in our analysis of the class-wise impact on the performance of sparse model transfer (\cref{sec:result_analysis}).
    \item $\mathcal{S}$ or ``\%-sparsity'' in all plots refers to the percentage of remaining weights in the model or, in other words, the capacity of the sparse model.  
    \item ``N-shots'' refers to the training data budget, i.e. the number of samples per class of the downstream dataset used during training.
\end{itemize}  

\subsection{Performance Analysis}

\subsubsection{Sparse Models}\label{sec:sparse_models}

\textbf{GMP:} \label{subsec:gmp}Analyzing the transfer performance of GMP-pruned \cite{zhu2017prune} ResNet-18 and ResNet-34 models at around a $80-90\%$ layer-wise sparsity level on various downstream datasets using ILM-VP, FLM-VP, and RLM-VP, we can see a clear detrimental impact of sparse models compared to their equivalent dense counterparts. 
\begin{figure}[!ht]
    \centering   \includegraphics[width=0.99\linewidth]{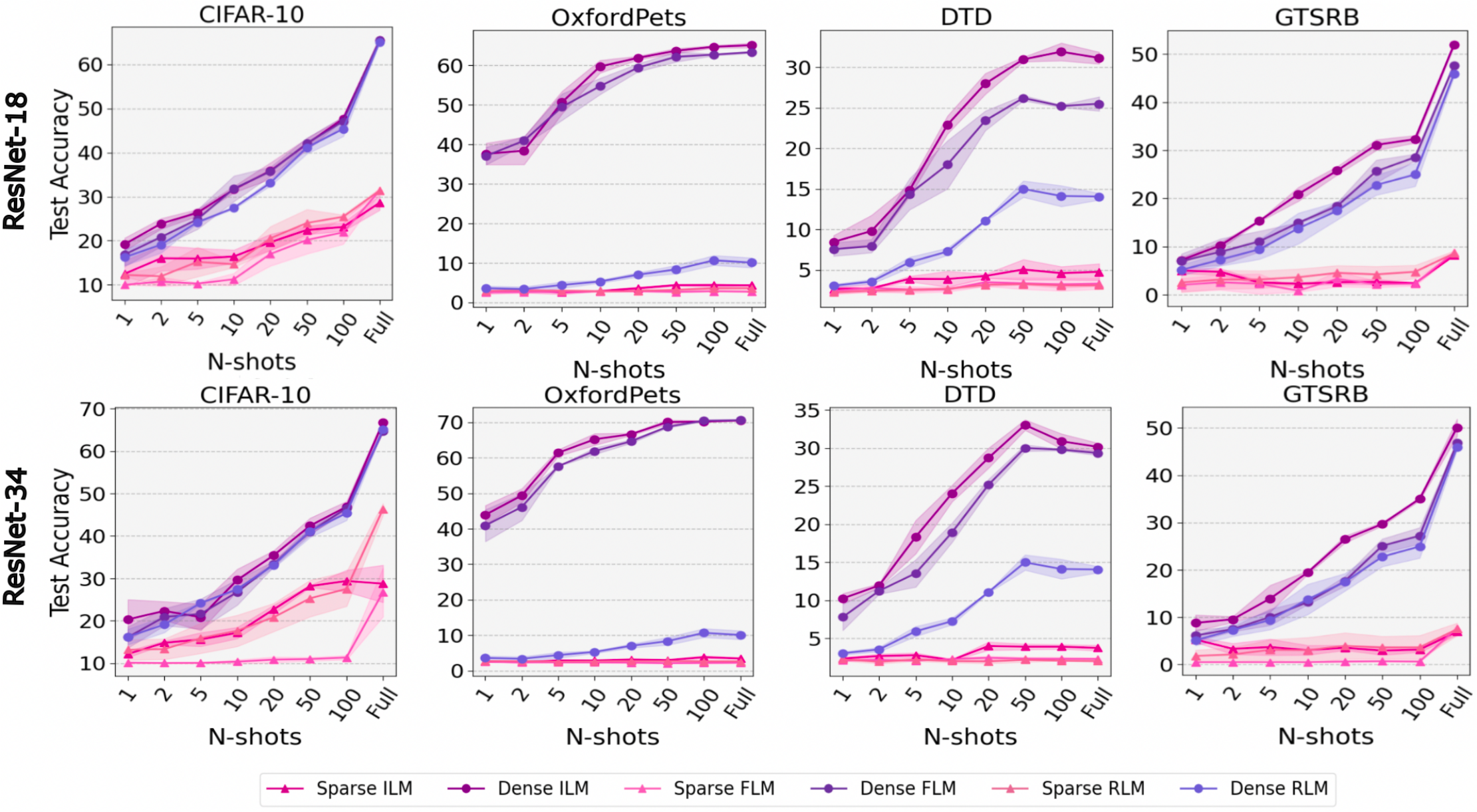}
    \caption{\textbf{GMP-pruned ResNet-18/34.} Transfer performance measured by test accuracy of pruned ResNet-18/34 model on a variety of downstream datasets and varying levels of data budgets.}
    \label{fig:resnet_gmp}
\end{figure} 

In Figure \ref{fig:resnet_gmp}, ResNet-18 analysis shows ILM-VP is generally the most effective for dense models, followed by FLM-VP. RLM-VP consistently lags, except in GTSRB where all VP methods perform similarly. Sparse models across settings consistently underperform dense counterparts, with no clear trend among different VP modes. Pruned model transfer impact is most significant in OxfordPets, with a consistent performance gap exceeding $50\%$ for various data budgets. Similarly, for ResNet-34 in Figure \ref{fig:resnet_gmp}, trends echo ResNet-18. Dense models outperform sparse counterparts across all data budgets and downstream datasets, with a notable accuracy improvement, especially in full data settings, perhaps because of the improved size of the model architecture compared to ResNet-18. ILM-VP remains consistently superior for both sparse and dense models.

In summary, we observe the detrimental impact of transfer via visual prompting methods on ResNet-18 and ResNet-34 models pruned using GMP across multiple downstream datasets and varying data budget settings, despite the models matching the dense model's upstream ImageNet-1k performance with $69.8\%$ for ResNet-18 and $73.3\%$ for ResNet-34.

\textbf{AC/DC and RigL:} 
We now study the transfer performance for pruned ResNet-50 models at $80\%, 90\%$ and $98\%$ sparsity compressed by AC/DC \cite{peste2021ac} and at $80\%, 90\%$ and $95\%$ sparsity compressed by RigL \cite{evci2020rigging}. 

\begin{figure}[!h]
    \centering   \includegraphics[width=0.99\linewidth]{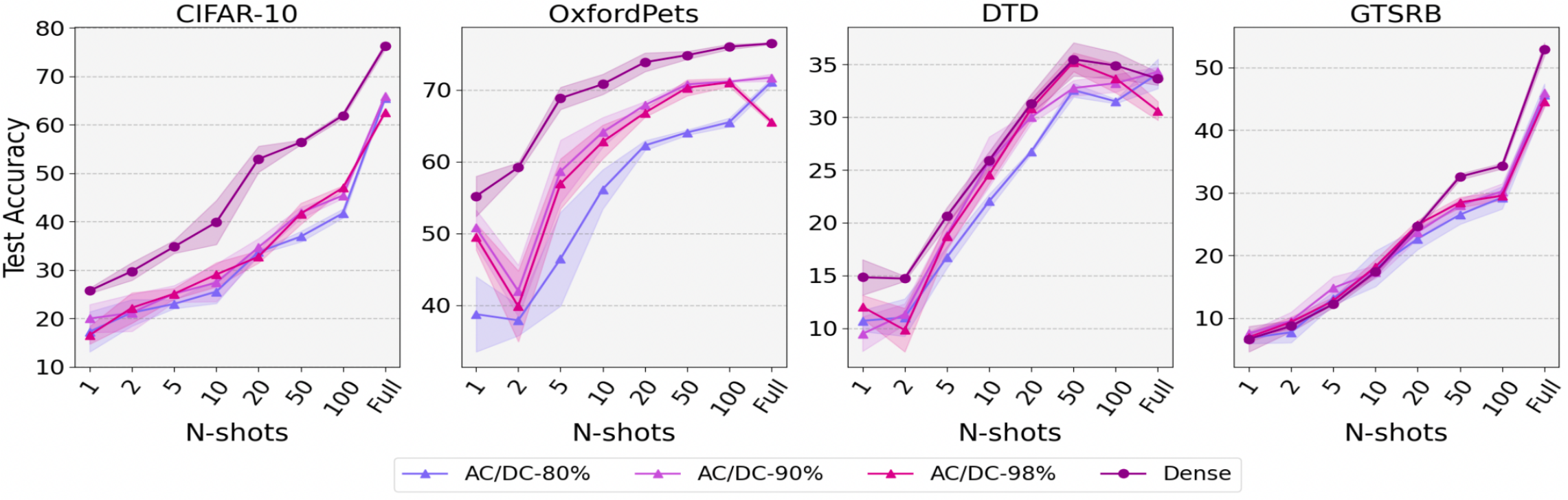}
    \caption{\textbf{AC/DC-pruned ResNet-50.} Transfer performance measured by test accuracy of pruned ResNet-50 model on a variety of downstream datasets and varying levels of data budgets.}
    \label{fig:acdc}
    \vspace{-2mm}
\end{figure}
\begin{figure}[!h]
    \centering   \includegraphics[width=0.99\linewidth]{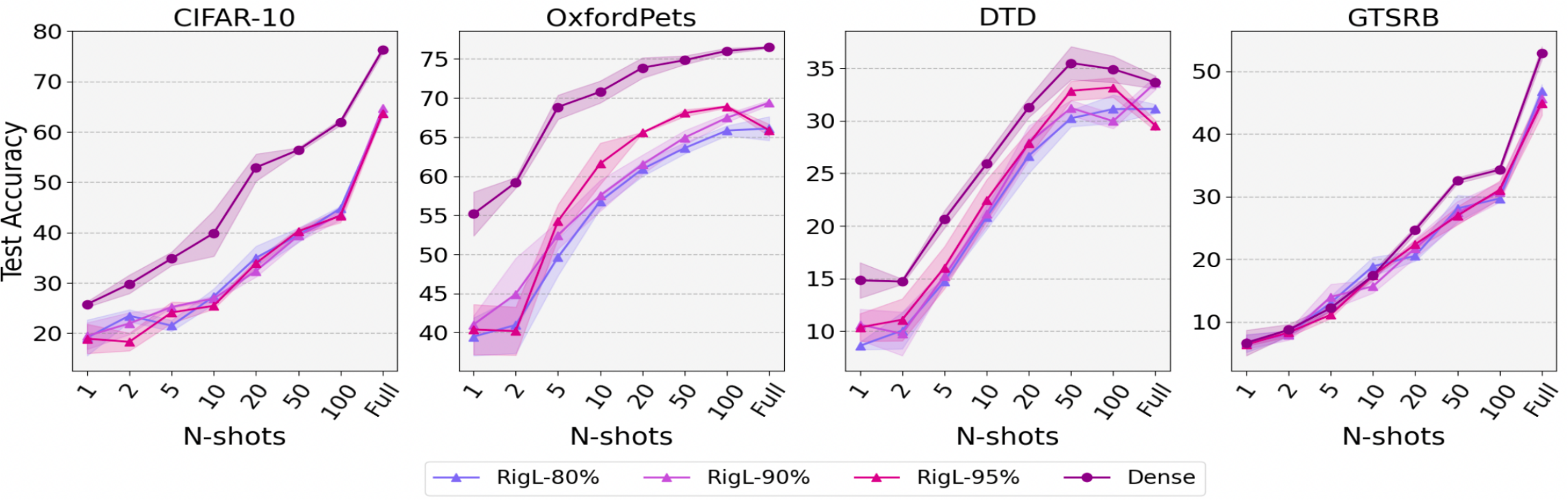}
    \caption{\textbf{RigL-pruned ResNet-50.} Transfer performance measured by test accuracy of pruned ResNet-50 model on a variety of downstream datasets and varying levels of data budgets.}
    \label{fig:rigl}
    \vspace{-2mm}
\end{figure}

In Figures \ref{fig:acdc} and \ref{fig:rigl}, the dense model generally outperforms sparse models for both AC/DC and RigL across four datasets. An exception is observed for GTSRB, where sparse models in specific sparsity and data budget settings match the performance of their dense counterpart. 
Comparing with the previously pruned ResNet-18 and 34 models using GMP, the ResNet-50 models pruned with AC/DC and RigL, while still generally worse than their dense counterparts, exhibit relatively better performance. The overall trends in performance on downstream tasks for models pruned by these dynamic sparsification techniques are quite similar.

We also examine the performance of Lottery Ticket Hypothesis (LTH) solutions for ResNet-50 when transferred at various sparsity configurations and data-budget settings in the supplementary material. Our conclusions primarily rely on the trends for ILM-VP, the sota method. In general, it is evident that the transfer of these LTH solutions using VP-based methods does not maintain their performance under low data volumes, despite their upstream performance matching or outperforming their dense counterparts.

\begin{table}[hb!]
    \small
    \resizebox{1.0\linewidth}{!}{
    \begin{tabular}{|cc|ccc|}
        \toprule
    \textbf{Datasets} &\textbf{Method} & \textbf{Dense (856.0M)} & \textbf{2x (473.7M)} & \textbf{4x (280.2M)} \\
        \midrule
     \multirow{2}{*}{\textbf{CIFAR-10}} & VP & \textbf{97.31 \%} &92.65 \% \textcolor{red}{(4.66 $\downarrow$)} & 90.07 \% \textcolor{red}{(7.24 $\downarrow$)} \\ 
        &ILM-VP & \textbf{97.53 \%} & 93.04 \% \textcolor{red}{(4.49 $\downarrow$)} &89.62 \% \textcolor{red}{(7.91 $\downarrow$)}\\
        \midrule

    \multirow{2}{*}{\textbf{Caltech101}} & VP & \textbf{96.26 \%} & 80.70 \% \textcolor{red}{(15.56 $\downarrow$)} & 73.90 \% \textcolor{red}{(22.36 $\downarrow$)} \\ 
        &ILM-VP & \textbf{95.45 \%} & 80.53 \% \textcolor{red}{(14.92 $\downarrow$)} &71.43 \% \textcolor{red}{(24.02 $\downarrow$)}\\
        \midrule

    \multirow{2}{*}{\textbf{OxfordPets}} & VP & \textbf{92.75 \%} & 60.86 \% \textcolor{red}{(31.89 $\downarrow$)} & 49.06 \% \textcolor{red}{(43.69 $\downarrow$)} \\ 
        &ILM-VP & \textbf{91.25 \%} & 58.79 \% \textcolor{red}{(32.46 $\downarrow$)} &46.17 \% \textcolor{red}{(45.08 $\downarrow$)}\\
        \midrule
        
    \multirow{2}{*}{\textbf{SVHN}} & VP & \textbf{95.06 \%} & 89.30 \% \textcolor{red}{(5.76 $\downarrow$)} & 89.21 \% \textcolor{red}{(5.85 $\downarrow$)} \\ 
        &ILM-VP & \textbf{94.51 \%} & 89.18 \% \textcolor{red}{(5.33 $\downarrow$)} &89.25 \% \textcolor{red}{(5.26 $\downarrow$)}\\
        \midrule
    \multirow{2}{*}{\textbf{GTSRB}} & VP & \textbf{91.43 \%}& 90.86  \% \textcolor{red}{(0.57 $\downarrow$)} & 90.73 \% \textcolor{red}{(0.70 $\downarrow$)} \\ 
        &ILM-VP & \textbf{91.06 \%} & 90.67 \% \textcolor{red}{(0.39 $\downarrow$)} &88.79 \% \textcolor{red}{(2.27 $\downarrow$)}\\
        \midrule
    \multirow{2}{*}{\textbf{DTD}} & VP &\textbf{54.36  \%}& 36.38  \% \textcolor{red}{(17.98 $\downarrow$)} & 31.70 \% \textcolor{red}{(22.66 $\downarrow$)} \\ 
        &ILM-VP & \textbf{54.04 \%} & 40.48 \% \textcolor{red}{(13.56 $\downarrow$)} &30.70 \% \textcolor{red}{(23.30 $\downarrow$)}\\
        \midrule
    \multirow{2}{*}{\textbf{EuroSAT}} & VP & \textbf{99.98 \%} & 98.03  \% \textcolor{red}{(1.95 $\downarrow$)} & 97.60 \% \textcolor{red}{(2.38 $\downarrow$)} \\ 
        &ILM-VP & \textbf{98.26 \%} & 97.78 \% \textcolor{red}{(0.98 $\downarrow$)} &97.36 \% \textcolor{red}{(0.90 $\downarrow$)}\\

        \bottomrule
    \end{tabular}%
    }
    \caption{\textbf{Visual Prompting on Compressed CLIP}. Performance comparison of VP \cite{bahng2022exploring} and ILM-VP \cite{chen2023understanding} on compressed CLIP ViT-L models across 7 datasets}
    \label{tab:res_clip}
\end{table} 

\subsubsection{Sparse Foundation Models}\label{sec:sparse_foundation_models}

This section presents the impact of visual prompting on compressed CLIP models. Specifically, we study the effect on transfer performance on compressed CLIP ViT-Large \cite{radford2021learning}, where we use two compressed models: 2x and 4x provided by UPop \cite{shi2023upop} for the image-retrieval task on the COCO dataset. UPop, a structured pruning framework for vision language transformers, adaptively allocates pruning ratios to selected model components and employs progressive pruning to achieve substantial compression ratios.

We present the results of Visual Prompting (VP) \cite{bahng2022exploring} and ILM-VP \cite{chen2023understanding} techniques applied to CLIP across seven datasets (see Table \ref{tab:res_clip}). We trained the visual prompt for 10 epochs with a batch size of 16 using an SGD optimizer and a cosine learning rate scheduler. Further details of the setup are available in the supplementary material. The table indicates that the transfer performance of the dense model is optimal compared to the compressed counterparts across all datasets, and the 4x compressed model exhibits the weakest performance. Although this effect is noticed to a lower extent on datasets like EuroSAT and GTSRB where the performance drop is only $\sim1$-$2\%,$ on datastes like Caltech101 and DTD this gap is much more pronounced with an average drop of $20\%$ and is exacerbated even further for OxfordPets where the transfer performance dips almost $45\%$ in the 4x compressed model. 

This outcome highlights how model compression negatively impacts not just the vision-only model transfer using visual prompting studied in the previous sections, but also the efficacy in downstream tasks in vision-language transformer-based models.

\begin{figure}[t]
\centering
    \small
\includegraphics[width=0.99\linewidth]{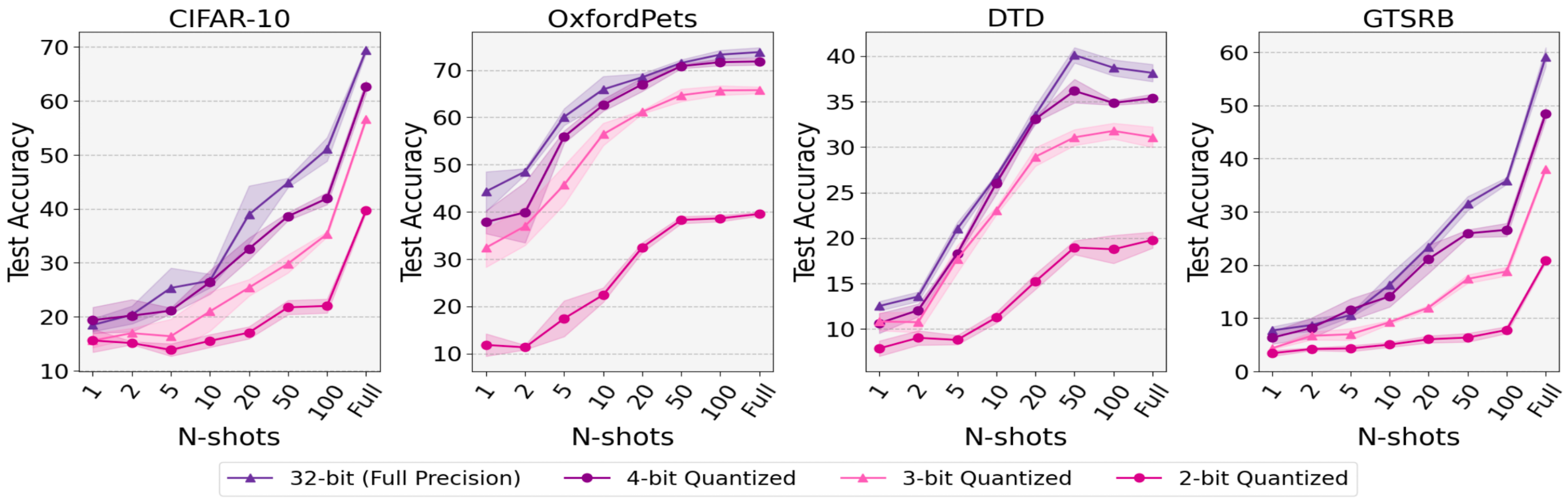}
    \caption{\textbf{VVTQuantized DeiT-T.} Transfer performance measured by test accuracy of quantized DeiT-T models on a variety of downstream datasets and varying levels of data budgets.}
    \vspace{-2mm}
    \label{fig:vvtq_deit}
\end{figure}

\begin{figure}[t]
\centering
    \small
\includegraphics[width=0.99\linewidth]{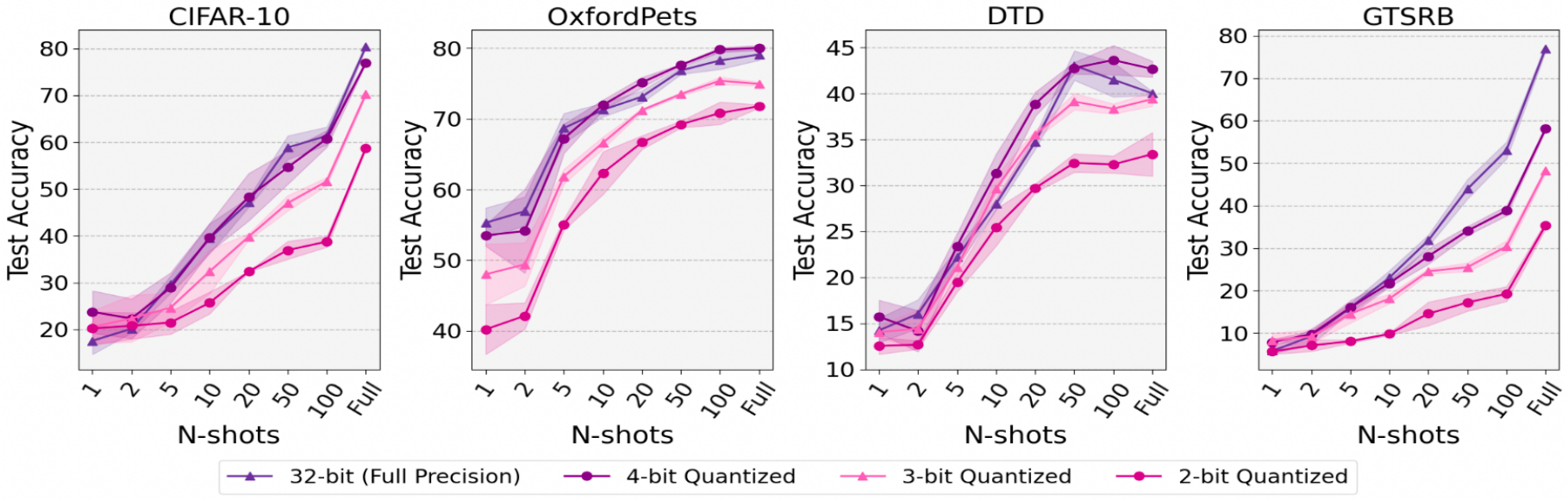}
    \caption{\textbf{VVTQuantized Swin-T.} Transfer performance measured by test accuracy of quantized Swin-T models on a variety of downstream datasets and varying levels of data budgets.}
    \vspace{-2mm}
    \label{fig:vvtq_swin}
\end{figure}

\subsubsection{Quantized Models}\label{sec:quantized_models}
We evaluate the performance of quantized Vision Transformer models, specifically comparing a full precision (32-bit) DeiT-T and Swin-T models with each of it's 4, 3, and 2-bit VVTQuantized DeiT-T and Swin-T model. \cite{huang2023variation}.

In Figure \ref{fig:vvtq_deit}, the overall trend indicates that the full precision DeiT-T model generally outperforms the quantized models across all datasets. Notably, at certain data budgets, the 4-bit DeiT-T model manages to match the performance of its full precision counterpart. The quantized models demonstrate clear performance differences as the quantization level increases. Similar trends are observed on additional datasets, as illustrated in the supplementary material. 

In Figure \ref{fig:vvtq_swin}, a similar trend is observed to that of the DeiT-T model, where the full precision model generally outperforms the quantized model. However, the gap between the 2-bit quantized model and others is smaller compared to what is seen in DeiT-T (more than 10\% in some datasets like OxfordPets and DTD). This suggests nuanced dynamics in the efficacy of quantization methods across different model architectures.

In comparison to the ResNet-50 models pruned with AC/DC and RigL in the previous section, we observe that the differences between a full precision and 4-bit quantized DeiT-T model are generally closer than those observed in sparsified ResNet-50 models. While sparsified models exhibit similar performance, in the quantization setting, as the level of quantization increases, the performance deviation among quantized models widens, especially for DeiT-T model. Quantized models and Specified AC/DC and RigL models demonstrate distinct behavioural deviations.
\subsubsection{Cross Modal Reprogramming Results}\label{sec:Cross Modal} In the Cross Modal Reprogramming setting, we extend our study to compare Dense Full Precision vision models against Sparse and Quantized vision models for NLP classification tasks.

The results in Table \ref{tab:reprogrammingexps} reveal that Dense models generally outperform Sparse models, with the Splice DNA dataset being the only exception where the Sparse model outperforms the Dense one. Intriguingly, the Full-Precision model and its 3-bits Quantized counterpart show similar performance across most tasks. We provide 4-bit and 2-bit results in Supplementary material which further support these findings. In cross-model reprogramming to NLP tasks, we observe consistent good performance across different bit precision quantization.
\setlength\tabcolsep{3pt}
\begin{table}[h]
\centering
\caption{\textbf{Cross Modal Reprogramming Accuracy.} Comparison of Dense Full Precision, Sparse, and Quantized Vision Models for NLP Classification Tasks.}\resizebox{0.98\linewidth}{!}{\begin{tabular}{@{}|l|cccc|cc|cc@{}}
\midrule
  \multicolumn{1}{|c|}{} & \multicolumn{2}{c|}{\emph{Dense}} & 
  \multicolumn{2}{c|}{\emph{Sparse}} & \multicolumn{2}{c|}{\emph{Quantized}}
  \\ \midrule
  Task & Resnet-34 & Resnet-50 & Resnet-34 & Resnet-50  & Deit 32-bit & Deit 3-bit
\\ \midrule
Yelp  & 91.01 $\pm$ 1.21 & 90.93 $\pm$ 1.07 & 87.58 $\pm$ 1.63 & 86.03 $\pm$ 3.21 & 91.83 $\pm$ 1.41 & 92.08 $\pm$ 1.48 \\
IMDB  &  79.90 $\pm$ 2.97 & 80.72 $\pm$ 3.75 & 68.46 $\pm$ 3.1 & 69.0 $\pm$ 0.52 & 77.86 $\pm$ 2.41 & 79.58 $\pm$ 2.64\\
\midrule
AG  & 91.01 $\pm$ 0.62	& 90.6 $\pm$ 0.86 & 84.23 $\pm$ 0.99 & 85.87 $\pm$ 1.77 & 91.67 $\pm$ 0.88 & 92.48 $\pm$ 0.99 \\
DBPedia  & 94.04 $\pm$ 1.50	& 94.12 $\pm$ 2.01 & 85.38 $\pm$ 0.86 &  78.19 $\pm$ 3.07	 & 96.90 $\pm$ 1.35 & 96.65 $\pm$ 1.26 \\
\midrule
Splice  &  80.96 $\pm$ 0.57	& 74.84 $\pm$ 7.11 & 90.69 $\pm$ 2.97 & 81.37 $\pm$ 3.63 & 78.35 $\pm$ 8.70 & 71.57 $\pm$ 3.8 	\\
\bottomrule
\multicolumn{1}{c}{}
\end{tabular}}
\label{tab:reprogrammingexps}
\end{table}
\setlength\tabcolsep{3pt}

We provide an expanded table of Cross Modal Reprogramming results for quantized models, now including 4 bit and 2 bit in Table \ref{tab:reprogrammingexps_supp}. The 4 bit and 3 bit quantized Deit models are generally able to keep similar performance to the full precision Deit model. The 2 bit model performs worse than the 32 bit model.

\begin{table}[h]
\centering
\caption{\textbf{Cross Modal Reprogramming Accuracy.} Comparison of Quantized Vision Models for NLP Classification Tasks.}\resizebox{0.98\linewidth}{!}{\begin{tabular}{@{}|l|cccc|}
\midrule
  \multicolumn{1}{|c|}{} & \multicolumn{4}{c|}{\emph{Quantized}}
  \\ \midrule
  Task & Deit 32-bit & Deit 4-bit & Deit 3-bit & Deit 2-bit
\\ \midrule
Yelp  & 91.83 $\pm$ 1.41 & 91.58 $\pm$ 1.21 & 92.08 $\pm$ 1.48 & 89.3 $\pm$ 2.6 \\
IMDB  & 77.86 $\pm$ 2.41& 78.68 $\pm$ 0.35 & 79.58 $\pm$ 2.64 & 79.0 $\pm$ 1.59 \\
\midrule
AG  &  91.67 $\pm$ 0.88 & 91.67 $\pm$ 1.47 &  92.48 $\pm$ 0.99& 89.34 $\pm$ 1.21 \\
DBPedia  & 96.90 $\pm$ 1.35 & - &  96.65 $\pm$ 1.26 &  95.18 $\pm$ 0.28 \\
\midrule
Splice  &  78.35 $\pm$ 8.70 & 76.96 $\pm$ 6.77 & 71.57 $\pm$ 3.8 & 61.76 $\pm$ 1.36	\\
\bottomrule
\multicolumn{1}{c}{}
\end{tabular}}
\label{tab:reprogrammingexps_supp}
\end{table}
%


\subsection{Calibration Analysis}\label{sec:calibration_analysis}

Although there has been an increase in studies that explore the performance \cite{iofinova2022well, linsparse, fedus2022review} and fairness \cite{tran2022pruning, hooker2019compressed, hooker2020characterising} of sparse models, only recently have studies been aimed at studying the calibration of pruned models \cite{arora2023quantifying, lei2023calibrating}. Yet, how calibration varies under different levels and methods of compression, such as pruning and quantization, is largely unexplored. This is also the case, especially for transfer via VP. To this end, in this section, we analyze the calibration trends of pruned and quantized models at varying levels of compression transferred by VP. Specifically, we inspect the expected calibration error (ECE) \cite{guo2017calibration} as a measure of transfer reliability under VP and how it varies with increasing compression levels. To the best of our knowledge, this is the first work to present an extensive study on the calibration of compressed models transferred using VP.

Although there have been novel formulations for calibration analysis that are suited for niche settings such as deep ensembles \cite{lakshminarayanan2017simple} and dropout training \cite{gal2016dropout}, the ECE remains an important measure of reliability and is appropriate for our relative comparisons. Furthermore, we only analyze calibration of models transferred under a full-data setting, as compressed models tend to overfit or `memorize' the training examples in a few-shot setting \cite{goodfellow2016deep, arpit2017closer}, thus producing predictions that are already overconfident, and a measure of reliability under these low data-volume settings would not be a representative comparison.

\begin{figure}[!h]
    \centering   \includegraphics[width=\linewidth]{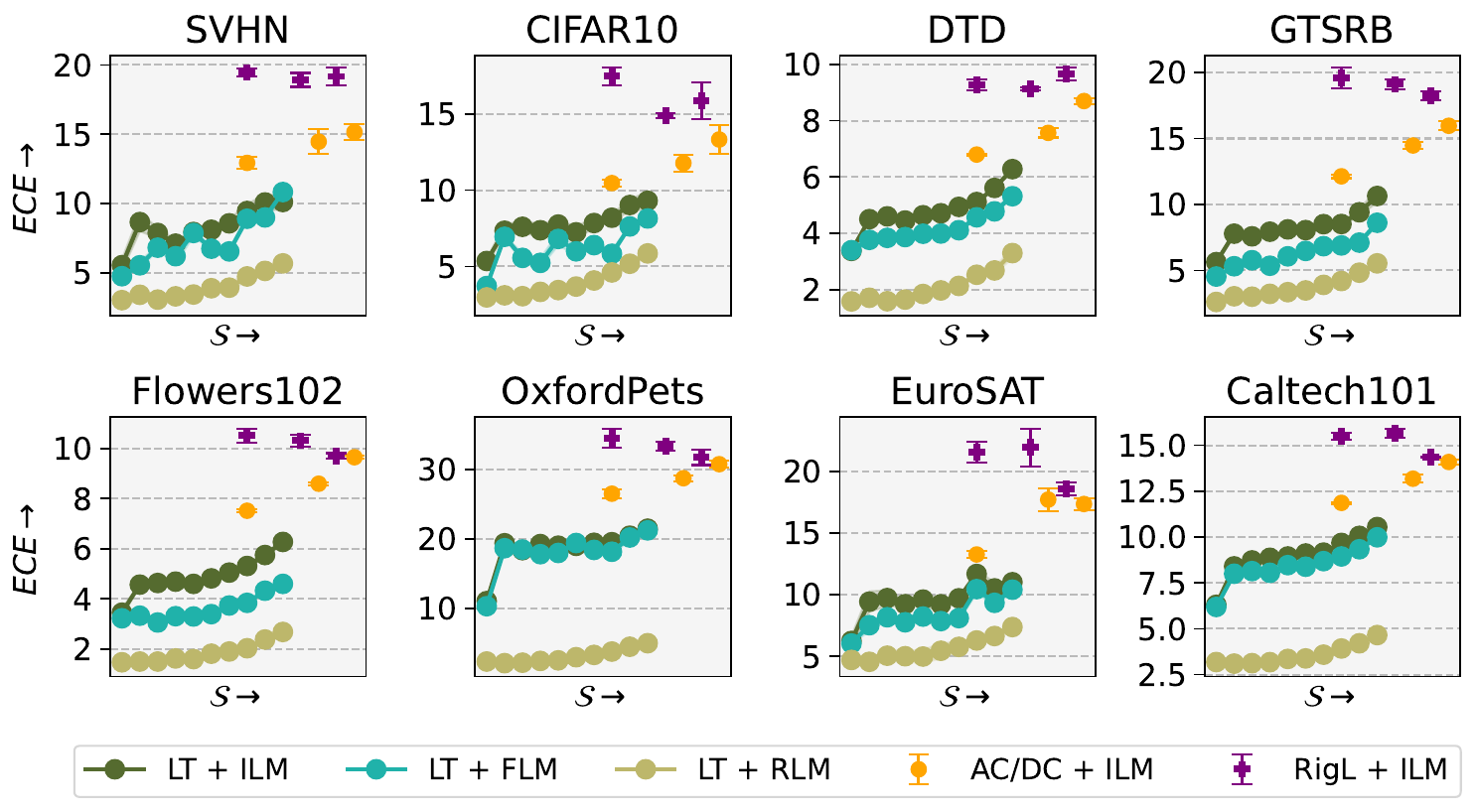}
    \caption{\textbf{Expected Calibration Error Analysis.} Comparison of ECE for LTH solutions of ResNet-50 models, and ResNet-50 models pruned by AC/DC and RigL transferred by ILM-VP across $8$ datasets measured against increasing levels of model sparsity, starting from dense (left) to sparsest (right). Lower ECE is better.}
    \label{fig:ece}
\end{figure} 

The formulation \cite{guo2017calibration} for ECE we use for our computations is given by,
\begin{align*}
    \textup{ECE} = \sum_{b=1}^{B} ~\dfrac{|M_b|}{N} ~\left|\textup{acc}(b) - \textup{conf}(b)\right|
\end{align*}
where $B$ is the total number of bins, $|M_b|$ is the number of predictions in bin $b$, $N$ is the total number of samples, and $\textup{acc}(b)$ and $\textup{conf}(b)$ are the accuracy and the confidence of bin $b$ respectively.

In Figure \ref{fig:ece}, we observe that the ECE of the dense transferred model (represented by the point on the left of each graph) is better than the sparse transferred models for all datasets, and this trend remains consistent regardless of the pruning technique used to compress the model.

Specifically for LTH solutions, the degradation in calibration performance on going from the dense model to the least-sparse transferred model is more pronounced for models that are transferred using ILM-VP and FLM-VP, while for RLM-VP we see a soft decline across most datasets. Furthermore, we see that with increasing levels of sparsity, the calibration of models continues to decline in all data-budget settings. The ECE for downstream performance on most datasets remains low around 10\%, with OxfordPets being a notable exception for which even the least sparse model has a much worse ECE on transfer, surpassing 20\%.

For models pruned by AC/DC and RigL and transferred by ILM-VP, we see that at the three levels of sparsity at which these models are evaluated, the ECE is significantly deteriorated compared to the dense counterpart. Note that the baseline for comparison for the AC/DC pruned models remains the dense ILM-VP calibration used for LTH since the underlying model remains ResNet-50 and these checkpoints are not progressively obtained. We also observe that the ECE of the AC/DC-pruned model is typically worse than that of the LTH solution when transferred using ILM-VP, and this effect is even more enhanced in the case of models pruned with RigL, which usually leads to models having relatively poorer calibration.

\begin{figure}[!h]
    \centering   \includegraphics[width=\linewidth]{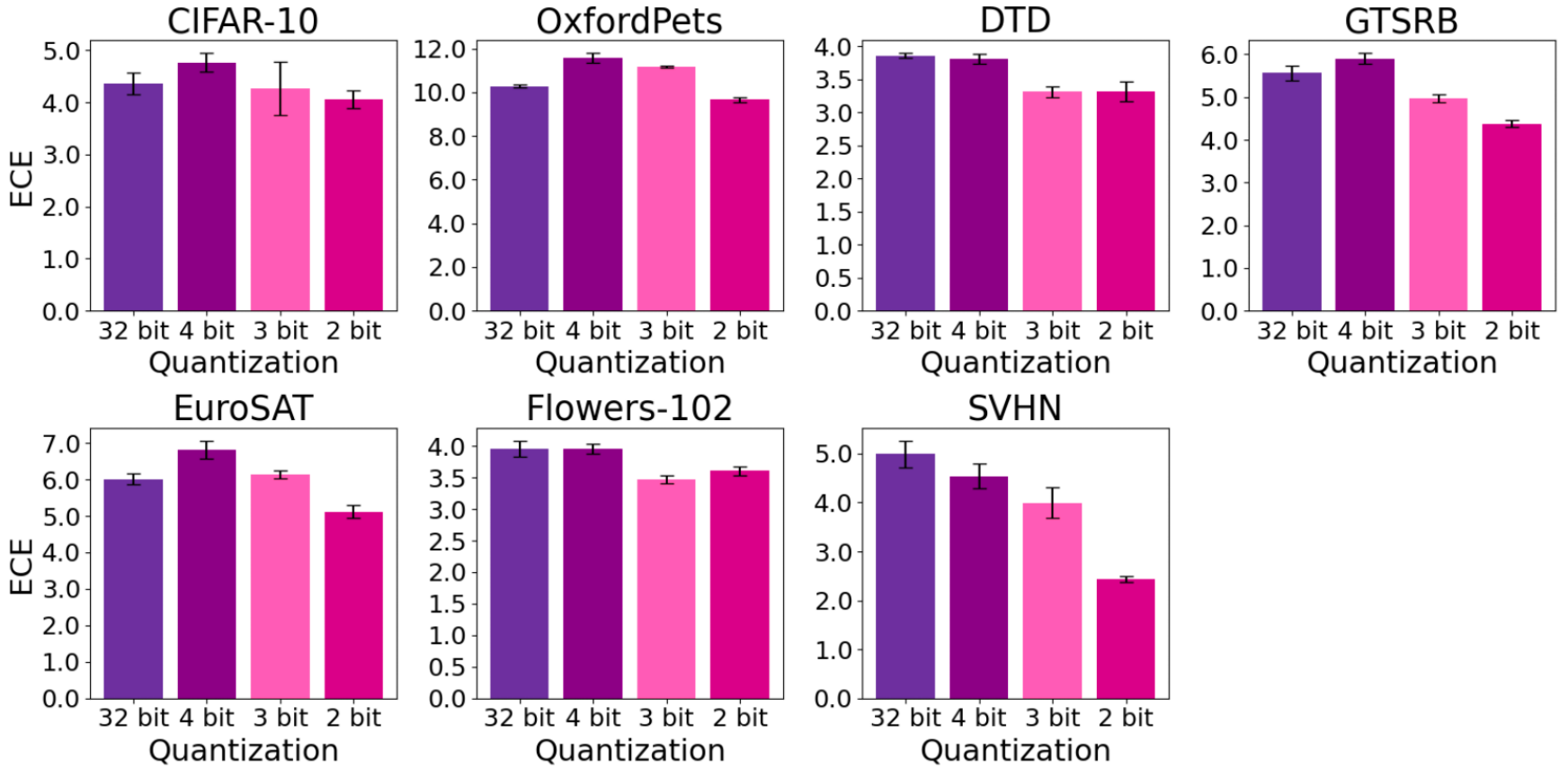}
    \caption{\textbf{DeiT ECE Analysis.} Comparison of ECE for DeiT quantized models across $7$ datasets measured against varying precision of quantization, starting from full-precision (32-bit) to 2-bit. Lower ECE is better.}
    \label{fig:deit_ece}
    \vspace{-6mm}
\end{figure} 

\begin{figure}[!h]
    \centering   \includegraphics[width=\linewidth]{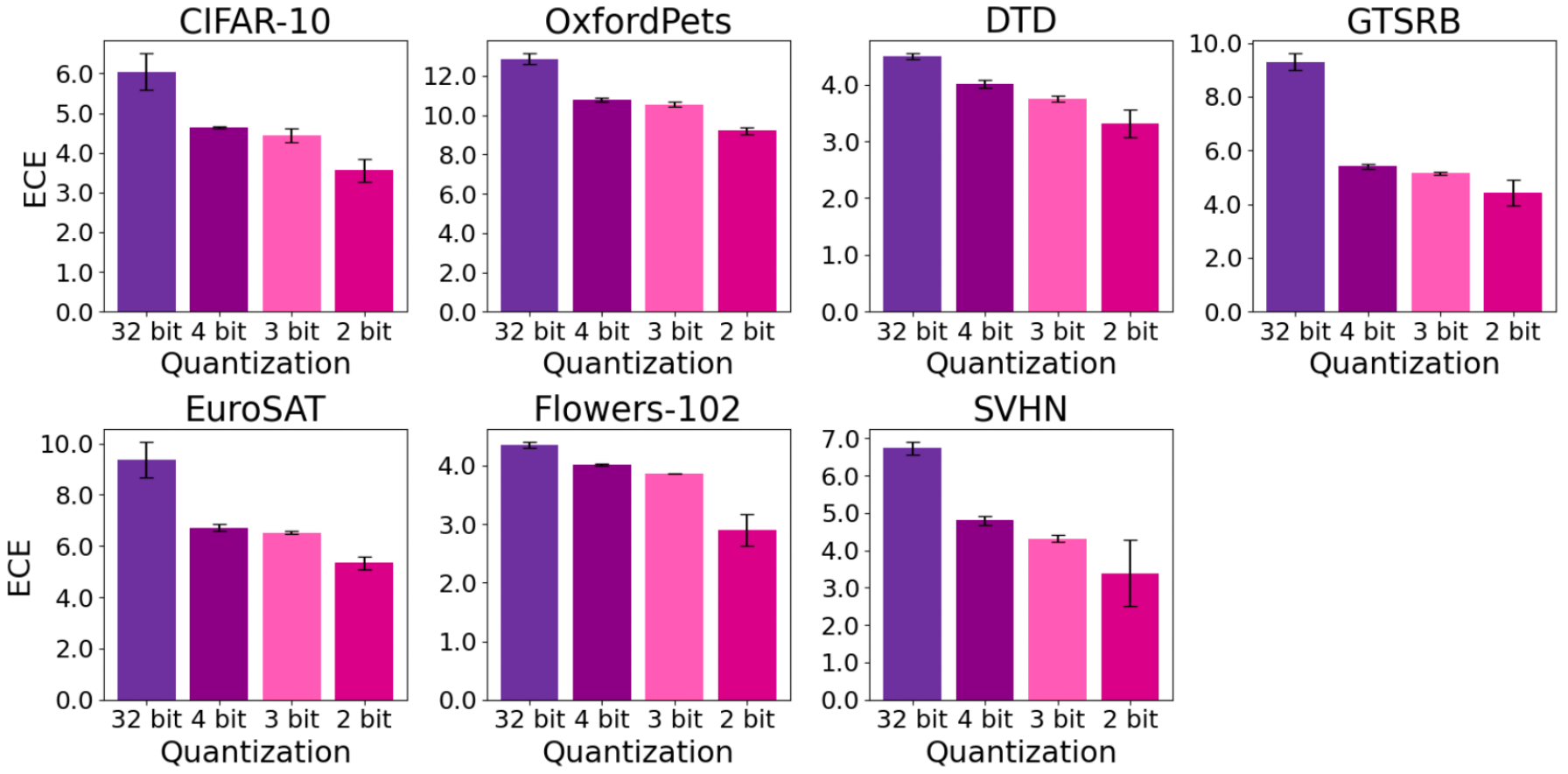}
    \caption{\textbf{Swin ECE Analysis.} Comparison of ECE for Swin quantized models across $7$ datasets measured against varying precision of quantization, starting from full-precision (32-bit) to 2-bit. Lower ECE is better.}
    \label{fig:swin_ece}
\end{figure} 

However, a distinct scenario unfolds when the compression method shifts to quantization. As depicted in Fig.~\ref{fig:deit_ece} and Fig.~\ref{fig:swin_ece}, an increasing quantization rate, transitioning from full-precision (32-bit) to 2-bit, consistently reduces or maintains the same Expected Calibration Error (ECE). This discrepancy underscores the disparity in calibration impact between models compressed via pruning and those compressed via quantization. To delve deeper into this phenomenon, we analyze the confidence values of both sparse and quantized models. Specifically, we examine two metrics: (a) Mean distance of the confidence distribution over classes for all incorrectly predicted samples compared to a uniform distribution, and (b) Mean confidence value for the correct class over all correctly predicted samples. We chose to measure the distance from the uniform distribution for incorrectly predicted samples because an ideal classifier should exhibit high uncertainty for incorrect predictions, indicating low confidence in assigning an incorrect label.

\begin{figure}[!h]
    \centering   \includegraphics[width=\linewidth]{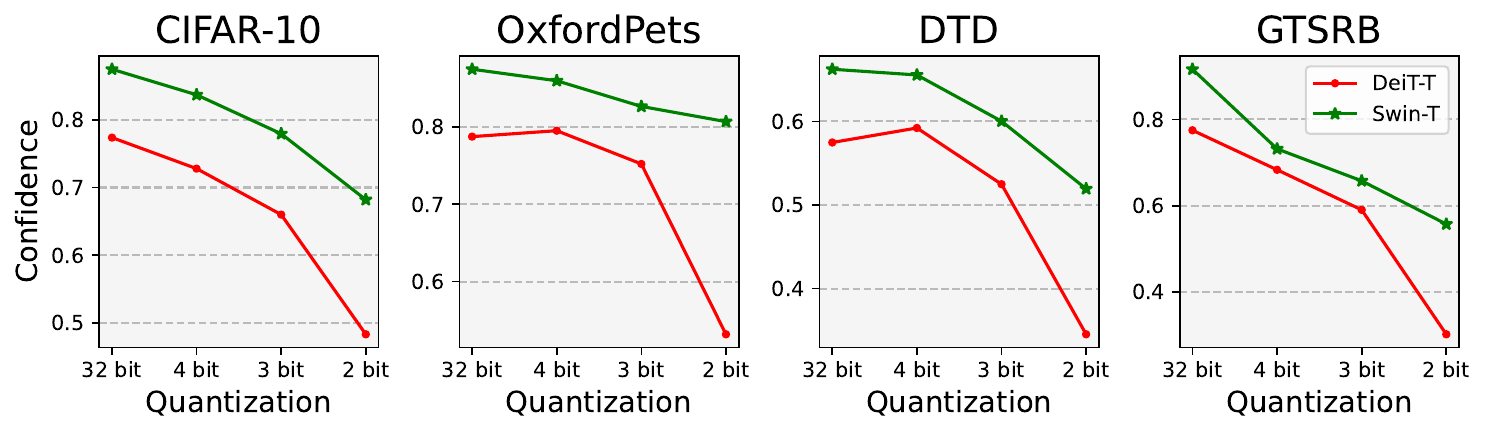}
    \caption{\textbf{Quantized Model Mean Confidence of Correct Prediction Analysis.} Observation of Mean confidence of the correct class for all accurate predictions for the DeiT-T and Swin-T models across full (32-bit) precision to 2-bit.}
    \label{fig:mean_correct_quant}
\end{figure} 

\begin{figure}[!h]
    \centering   \includegraphics[width=\linewidth]{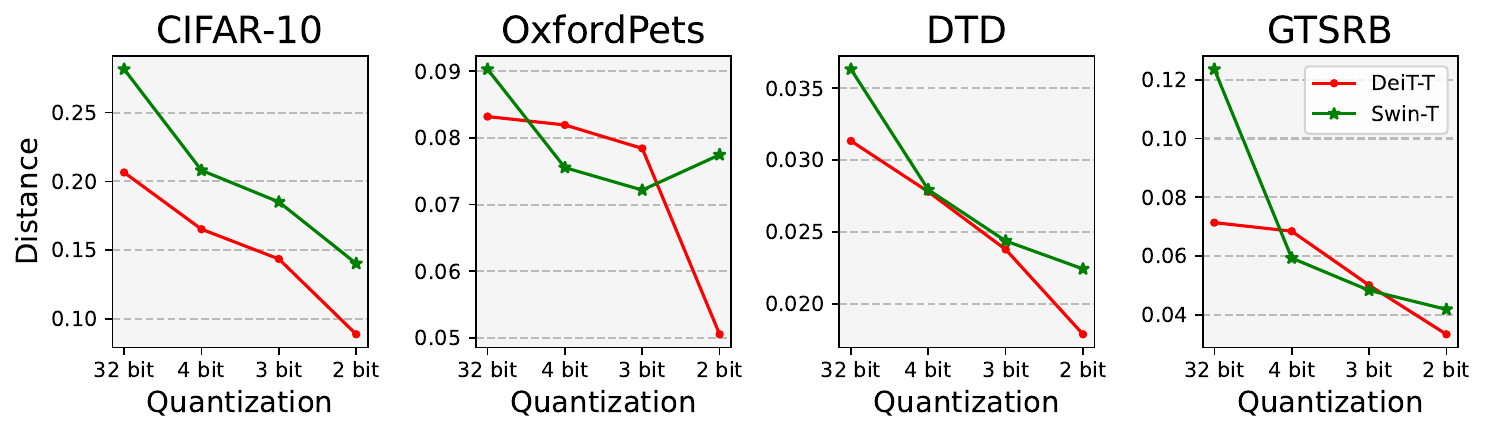}
    \caption{\textbf{Quantized Model Mean Distance to Uniform Distribution Analysis.} Observation of Mean of KL Divergence from the confidence distribution of incorrect model predictions to the uniform distribution for DeiT-T and Swin-T models across full (32-bit) precision to 2-bit.}
    \label{fig:mean_dist_incorrect_quant}
\end{figure} 

In the case of quantized models, as depicted in Fig.~\ref{fig:mean_correct_quant} and Fig.~\ref{fig:mean_dist_incorrect_quant}, we observe an optimal pattern: as accuracy declines from the full-precision (32-bit) to the 2-bit variant, the mean confidence of correctly predicted class labels consistently decreases. Simultaneously, the mean distance of the confidence distribution from the uniform distribution for incorrectly classified samples also decreases monotonically, indicating a higher level of uncertainty when the overall model performance deteriorates.

\begin{figure}[!h]
    \centering   \includegraphics[width=\linewidth]{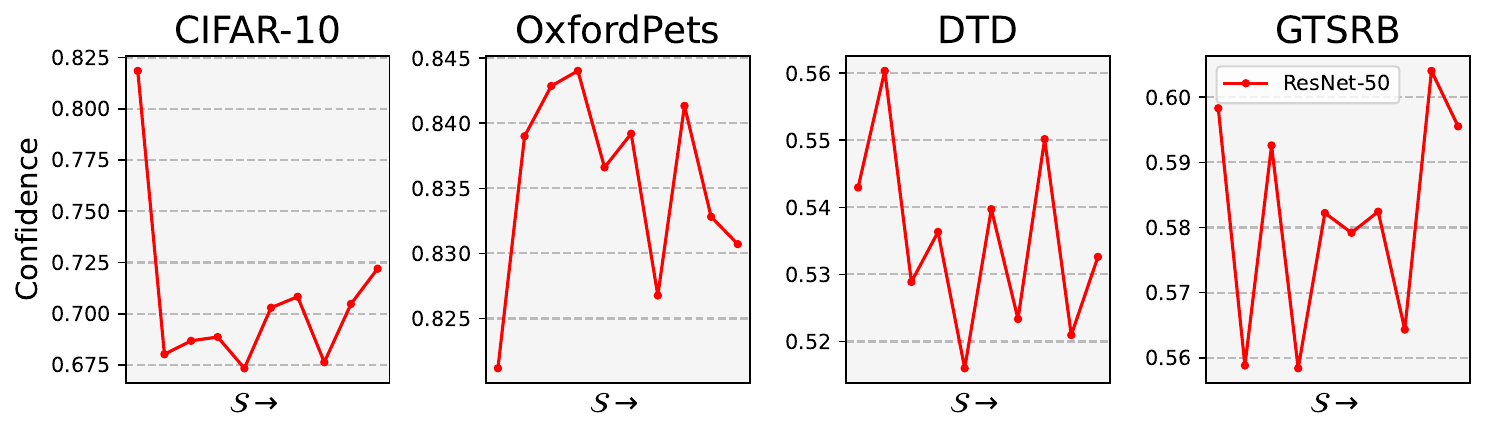}
    \caption{\textbf{Sparse Model Mean Confidence of Correct Prediction Analysis.} Mean Confidence of Correct Class for accurately predicted samples via ResNet-50 lottery ticket sparse models at different levels of sparsity for varying downstream target datasets. All results were analyzed via ILM-VP mode of transfer.}
    \label{fig:mean_correct_sparse}
\end{figure} 

\begin{figure}[!h]
    \centering   \includegraphics[width=\linewidth]{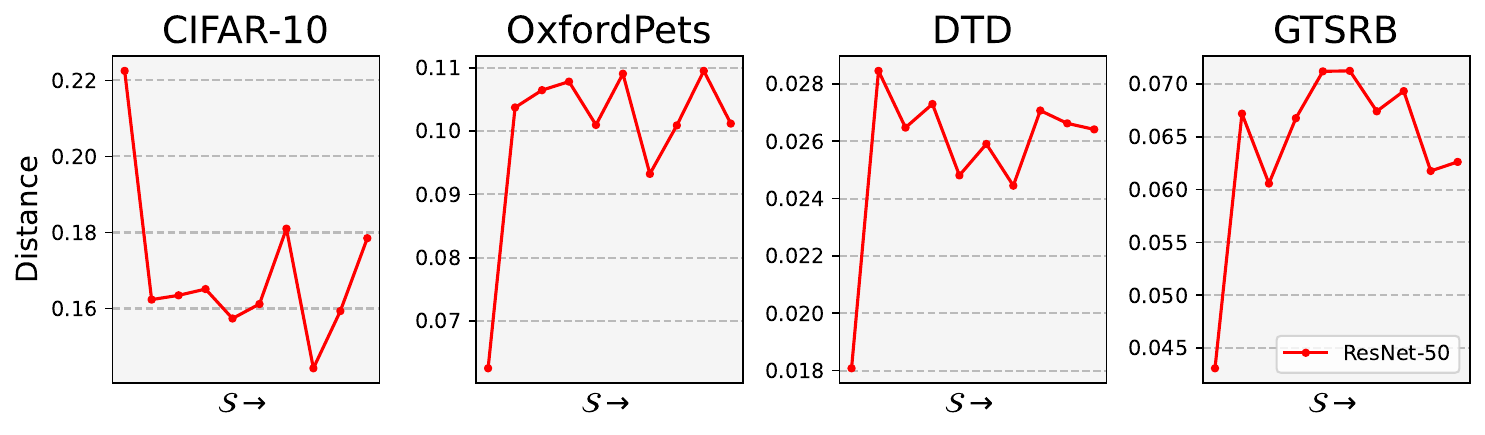}
    \caption{\textbf{Sparse Model Mean Distance to Uniform Distribution Analysis.} Mean of KL Divergence Distance between the confidence distribution of incorrectly classified samples and uniform distribution for sparse lottery ticket variants of ResNet-50 at varying levels of sparsity. All results were analyzed via ILM-VP mode of transfer.}
    \label{fig:mean_dist_incorrect_sparse}
\end{figure} 

However, when considering sparse models (LT), the optimal trend observed in quantized variants no longer persists. Specifically, as illustrated in Fig.~\ref{fig:mean_correct_sparse}, the confidence associated with the correct class label for accurately predicted samples fluctuates with an increasing rate of sparsity. More significantly, as sparsity intensifies, the distance of the confidence distribution for incorrectly classified samples from the uniform distribution either rises or fluctuates around the original dense model's values, indicating a higher level of overconfidence.

Given that Expected Calibration Error (ECE) is a metric representing the mean absolute difference between accuracy and corresponding confidence values, the observed trend in confidence values for sparse models elucidates why they encounter an increase in ECE. On the contrary, the more optimal trend observed in the case of quantization generally results in a lower or comparable ECE to that of the full dense, full-precision model.

\section{Conclusion}\label{sec:conclusion}
In this work, we present an extensive study on the transfer of models using visual prompting methods on downstream classification tasks on the axis of model compression and low data volume. Our findings on a large number of datasets using pruned and quantized models from various vision-based architectures, as well as vision-language transformer-based architectures suggest the existence and universality of hidden cost in downstream performance drop when using visual prompting. We further show the detrimental impact of model sparsity on the calibration of the transferred model across models pruned by various techniques, which usually worsens with an increasing level of sparsity. Furthermore, on the contrary, we demonstrate that quantization as a method of compression does not exhibit the same negative impact on calibration as attributed in the case of sparse (pruned) models. With the rapid advances in vision(-language) foundation models, visual prompting can be a crucial technique for downstream task adaptation, similar to the indispensable role of text prompting for large-language models, and our results provide important insights and motivations into the future design of prompting-friendly model compression methods. Following our analysis into the hidden costs associated with model compression, our aim is to extend our work by using influence estimation \cite{feldman2020neural} to characterize the positive or negative contribution of certain data sub-populations to transfer performance under the visual prompting regime, and also on a wider range of compression techniques.
\clearpage \newpage
{
    \small
    \bibliographystyle{ieeenat_fullname}
    \bibliography{main}
}

\clearpage
\setcounter{page}{1}
\maketitlesupplementary

\section{Background on Model Reprogramming}

Mathematically, let $\{\mathbf{x}_i,\mathbf{y}_i\}_{i=1}^n$ denote $n$ pairs of data samples $\{\mathbf{x}_i\}$ and their classification labels $\{\mathbf{y}_i \}$ for a target image classification task.
$\mathbf{x}_i \in \mathbb{R}^{w \times h \times c}$ and $w$, $h$, $c$ are the image width, height, and number of color channels,  respectively.  $\mathbf{y}_i \in \{1,2,\ldots,K\}$ and $K$ is the total number of image class labels. Let $f_\theta(\cdot)$ denote a pre-trained image classifier parametrized by $\theta$, which takes an image $\mathbf{x} \in \mathbb{R}^{w' \times h' \times c}$ as input and gives a prediction $f_\theta(\mathbf{x})$ of prediction probabilities over $K'$ classes, where $K \leq K'$, $w \leq w'$, and $h \leq h'$. In the standard VP training procedure, a masked perturbation, denoted by $\mathbf{M} \odot \mathbf{\delta} $, is appended to a zero-padded version of  $\{\mathbf{x}_i\}$ (denoted by $\{\mathbf{x'}_i\}$) in order to match the input dimension of the pre-trained model.  The binary mask $\mathbf{M} \in \{1,0\}^{w \times h \times c}$ denotes where to add the trainable perturbation to zero-padded images, and $\mathbf{\delta} \in \mathbb{R}^{w \times h \times c}$ serves as a trainable universal perturbation. At the model output, a mapping function $h_k$ is assigned for every target class label $k \in \{1,2,\ldots,K\}$ such that $h_k(f_\theta(\mathbf{x'}+\mathbf{M} \odot \mathbf{\delta} ))$ gives the prediction probability of the class $k$ for an image $\mathbf{x}$ in the target domain. Finally, VP trains the parameters associated with the input transformation (e.g. $\mathbf{\delta}$) and/or the output mapping layers (e.g. if $\{h_k\}_{k=1}^K$ has trainable parameters) based on task-specific loss evaluated on $\{\mathbf{x}_i,\mathbf{y}_i\}_{i=1}^n$.

\section{Experiment Details}\label{sec:experiment_app}

In this section, we describe the hyperparameters used for all the experiments shown in \cref{sec:sparse_models} in the main text and the experiments in the appendix.

\begin{table}[h]
  \centering
  \begin{tabularx}{\linewidth}{>{\centering\arraybackslash}X|*{8}{>{\centering\arraybackslash}X}}
    \textbf{NS} & 1 & 2 & 5 & 10 & 20 & 50 & 100 & Full \\
    \hline
    \textbf{BS} & 8 & 8 & 16 & 32 & 32 & 64 & 128 & 128 \\
  \end{tabularx}
  \caption{The different batch sizes (BS) used for each N-shots (NS) configuration for all experiments in \cref{sec:sparse_models} and \cref{sec:lth_appendix}}
  \label{tab:batch_size}
\end{table}

For all experiments discussed in \cref{sec:sparse_models} and \ref{sec:lth_appendix}, we train the model for 100 epochs using the Adam optimizer \cite{kingma2014adam}. We employ a multistep learning rate decay scheduler that reduces the learning rate by a factor of $\frac{1}{10}$ at the 50th and 72nd epochs, respectively, from its initial value of 0.01. The specific batch sizes used for training with each N-shots configuration are detailed in Table \ref{tab:batch_size}. To ensure reproducibility, we utilized checkpoints from \cite{iofinova2022well} for the RigL \cite{evci2020rigging} and AC/DC \cite{peste2021ac} models, while the sparse checkpoints for ResNet-18, ResNet-34 and all the VGG \cite{simonyan2014deep} variants were obtained from \href{https://sparsezoo.neuralmagic.com/}{NeuralMagic SparseZoo}. The dense checkpoints used were imported from the Torchvision library \cite{torchvision2016}.

In subsequent sections, we present additional experiments that go beyond the scope of the results outlined in the main text.

\section{Additional LTH Experiments}\label{sec:lth_appendix}

\textbf{Lottery Ticket Hypothesis:}\label{sec:lottery_ticket_hypothesis} In this section, we study the performance of LTH solutions for ResNet-50 when transferred at various configurations of sparsity states at different data-budget settings. In the left subplot for each dataset, we report the test accuracy of the dense model and the sparsest model (at $\sim$12\% sparsity) for transfer via the three VP methods, while each cell of the heatmap subplots on the right represents the mean difference across seeds between the dense model performance versus the sparse model transfer at a specified state of $(\text{sparsity}, \text{N-shot})$ pair. We only show the heatmaps for transfer via ILM-VP and FLM-VP in the main text and reflect on the RLM-VP heatmaps in the Supplement. \cref{sec:lth_appendix} due to the generally poor performance of RLM-VP regardless of model sparsity or low data volumes compared to the other two VP methods. The upstream performance of the LTH solutions at different sparsity levels used in this study is shown in Figure \ref{fig:LT}. We show here the results on CIFAR-10, OxfordPets, DTD, and Caltech101 while the others are in supplementary \cref{sec:lth_appendix}.

\begin{figure}[!h]
    \centering   \includegraphics[width=0.9\linewidth]{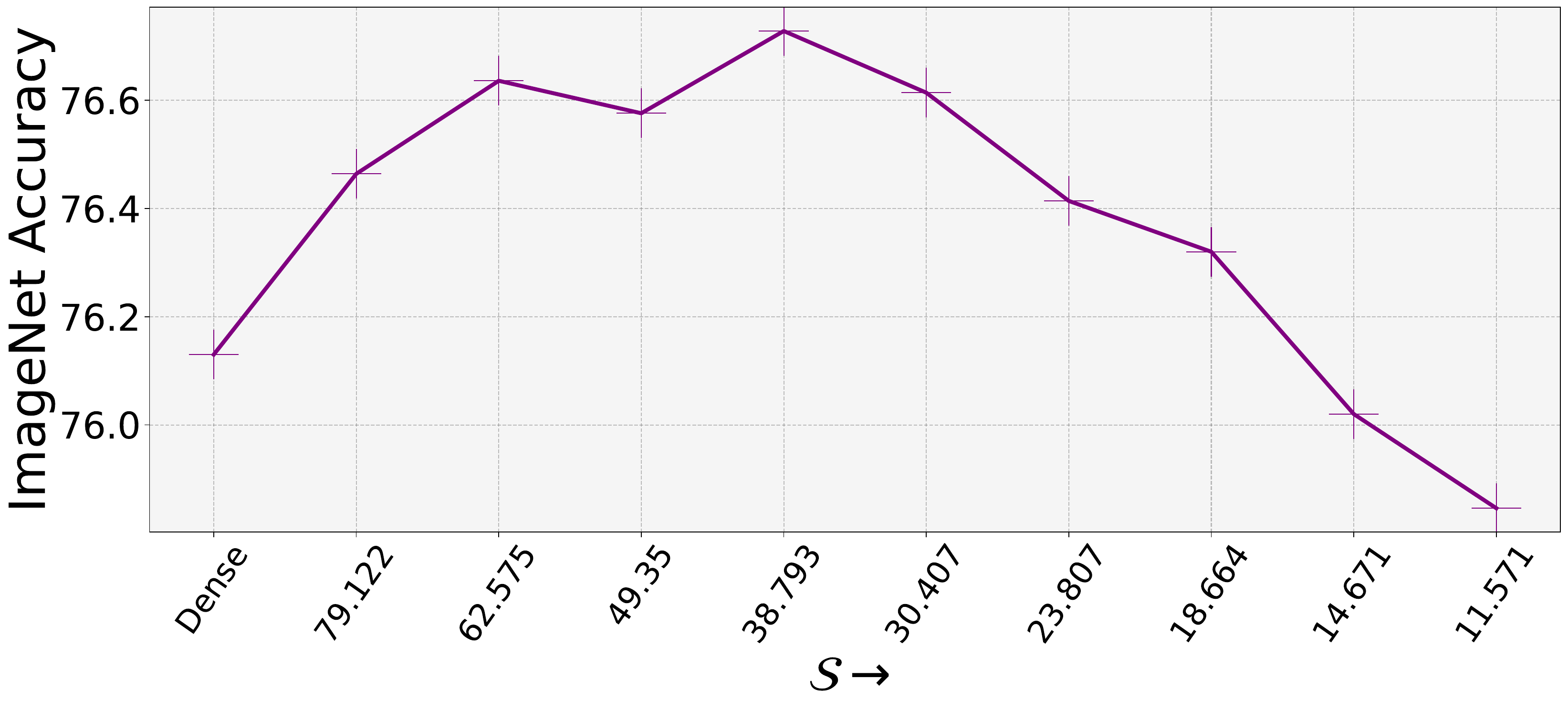}
    \caption{Top-1 accuracy of \textbf{LTH solutions} of a ResNet-50 pre-trained on the ImageNet-1k at different sparsity ($\mathcal{S}$) states.}
    \label{fig:LT}
    \vspace{-2mm}
\end{figure}

\begin{figure}[!h]
    \centering   \includegraphics[width=0.99\linewidth]{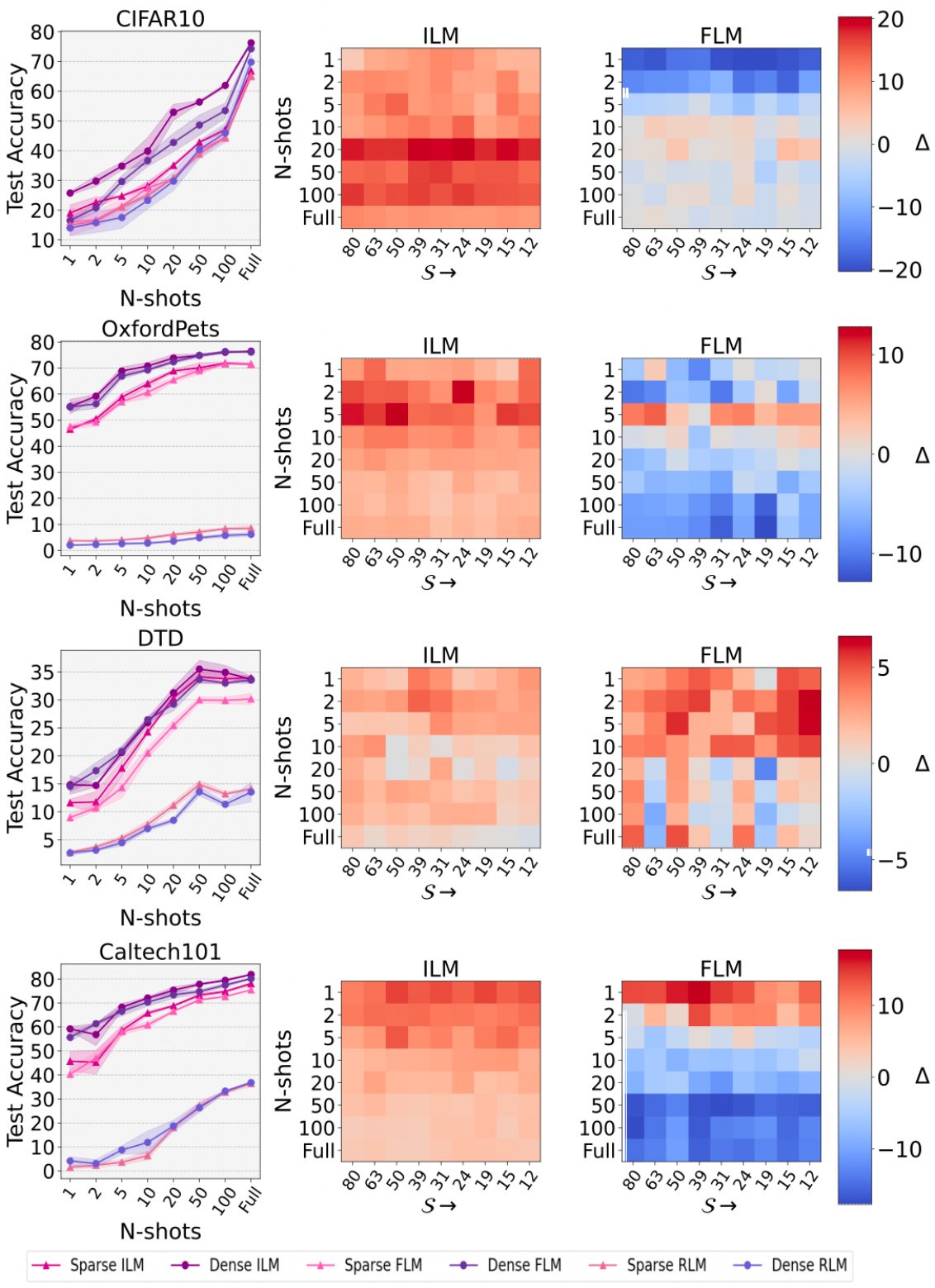}
    \caption{\textbf{Performance Gap of LT solutions on Transfer.} In each subfigure, from left to right, the first subplot represents the comparison between transfer via various VP methods for both the dense network and LT at $\approx$ 11\% sparsity on CIFAR-10 (\textbf{top}), OxfordPets, DTD and Caltech101 (\textbf{bottom}) at different target $N$-shot settings, while the two subplots on the right represent the $\Delta-$difference for ILM-VP and FLM-VP respectively. $\mathcal{S}$ represents increasing levels of \%-sparsity levels of Dense (leftmost), $79.122, ~62.575, ~49.35, ~38.793, ~30.407$, $23.807, ~18.664, ~14.671,$ and $11.571$ (rightmost).}
    \label{fig:lth_main_results}
    \vspace{-2mm}
\end{figure}

Taking CIFAR-10 as an example, from the test-accuracy subplot in Fig. \ref{fig:lth_main_results} (\textbf{top}) that for every N-shots budget, the dense model performed superior to that of the LT for transfer via ILM-VP and FLM-VP, and this trend holds true for all other datasets as well. Furthermore, we observe that ILM-VP outperforms FLM-VP in all N-shot settings. RLM-VP, which generally has a much lower performance compared to other VP methods, shows a slight deviation from this trend where we see that the sparse model tends to match the performance of the dense model, especially at transfer settings with a higher data budget.

Furthermore, we observe that the detrimental impact on performance due to LTs was significantly more pronounced in the case of ILM-VP where, for example, in the case of 20-shot configurations, LTs in all sparsity states studied in this work had on average a 20\% reduction in top-1 accuracy compared to their dense counterpart (see Figure \ref{fig:lth_main_results} (\textbf{top})). In the case of FLM-VP, the performance of LTs was actually better than that of the dense model for the few-shot settings as seen in the case of the one-shot and two-shot data budget settings; however, at higher data budget settings, the degradation of performance increases. The trend for ILM-VP transfer remains consistent across all four datasets, but there is a variation in that of FLM-VP based transfer. Specifically, for OxfordPets while the LTs overall seem to match or outperform their dense counterparts, for Caltech101 (see Figure \ref{fig:lth_main_results} (\textbf{bottom})) this only holds for the higher data budget settings. For DTD, on the other hand, barring a few data budget settings, the sparse model transfer seems to hurt the transfer performance overall.

We primarily base our conclusions on the trends for ILM-VP as it is the SOTA method, and thus in general, it is clear that the transfer of these LT solutions using VP-based methods does not keep their performance intact under low data volumes, although their upstream performance matches or outperforms their dense counterpart (see Figure \ref{fig:LT}).

In this section, we expand upon the findings of the lottery ticket hypothesis (LTH) discussed in Section \cref{sec:lottery_ticket_hypothesis}. We present the results of transferring ResNet-50 LTH solutions using ILM-VP \cite{chen2023understanding} and FLM-VP \cite{tsai2020transfer} to the four remaining downstream datasets: SVHN \cite{netzer2011reading}, GTSRB \cite{Houben-IJCNN-2013}, Flowers102 \cite{Nilsback08}, and EuroSAT \cite{helber2018introducing}. Subsequently, we evaluate the transfer performance of RLM-VP on all eight datasets.

\begin{figure}[!ht]
    \centering   \includegraphics[width=0.99\linewidth]{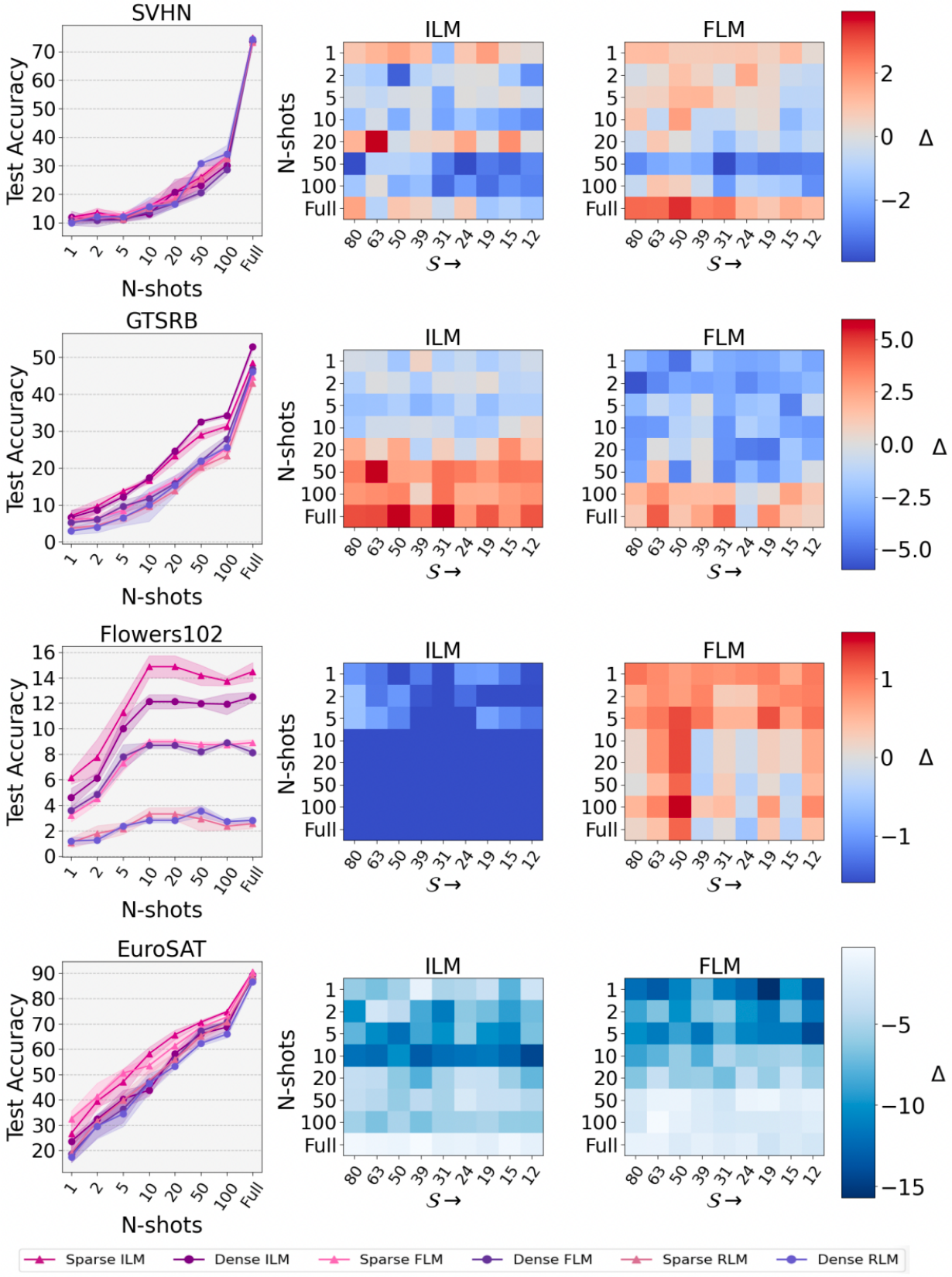}
    \caption{\textbf{Performance Gap of LT solutions on Transfer.} In each subfigure, from left to right, the first subplot represents the comparison between transfer via various VP methods for both the dense network and LT at $\approx$ 11\% sparsity on SVHN (\textbf{top}), GTSRB, Flowers102, and EuroSAT (\textbf{bottom}) at different target $N$-shot settings, while the two subplots on the right represent the $\Delta-$difference for ILM-VP and FLM-VP respectively. $\mathcal{S}$ represents increasing levels of \%-sparsity levels of Dense (leftmost), $79.122, ~62.575, ~49.35, ~38.793, ~30.407$, $23.807, ~18.664, ~14.671,$ and $11.571$ (rightmost).}
    \label{fig:lth_appendix_results}
    \vspace{-2mm}
\end{figure}

From the heat maps depicted in Figure \ref{fig:lth_appendix_results}, a notable contrast emerges when comparing the performance on SVHN, GTSRB, and Flowers102 with the four datasets discussed in \cref{sec:lottery_ticket_hypothesis} in the main text. On these three datasets, the discrepancy between the sparse and dense models is relatively small, generally within the range of approximately 5\%.

For SVHN, it is observed that sparse model transfer yields a performance dip compared to its dense counterpart in sporadic instances of low- and high-data volumes, particularly noticeable in the case of FLM-VP. However, in other data volume scenarios, the sparse model either matches the dense model or slightly outperforms it, regardless of the visual prompting (VP) method employed.

In the case of GTSRB, under higher data volumes, transferring the sparse model results in performance degradation compared to the dense models for both ILM-VP and FLM-VP. Conversely, for Flowers102, while the transfer of sparse models through FLM-VP incurs performance loss across nearly all sparsity and data volume settings, ILM-VP exhibits a reversed trend. Here, sparse models outperform the dense counterpart by approximately 1\% in high data volume settings and by around 0-1\% in lower data volume settings.

For EuroSAT, it is observed that, for both FLM-VP and ILM-VP, the sparse model outperforms the dense counterpart, particularly in low-data-volume scenarios.

The accompanying test accuracy plot (leftmost) in Figure \ref{fig:lth_appendix_results} reveals that, except for Flowers102, the performance of sparse and dense model transfer under all three visual prompting methods is closely aligned, with ILM-VP often exhibiting a slight edge over the other two methods.

\begin{figure}[!ht]
    \centering   \includegraphics[width=0.99\linewidth]{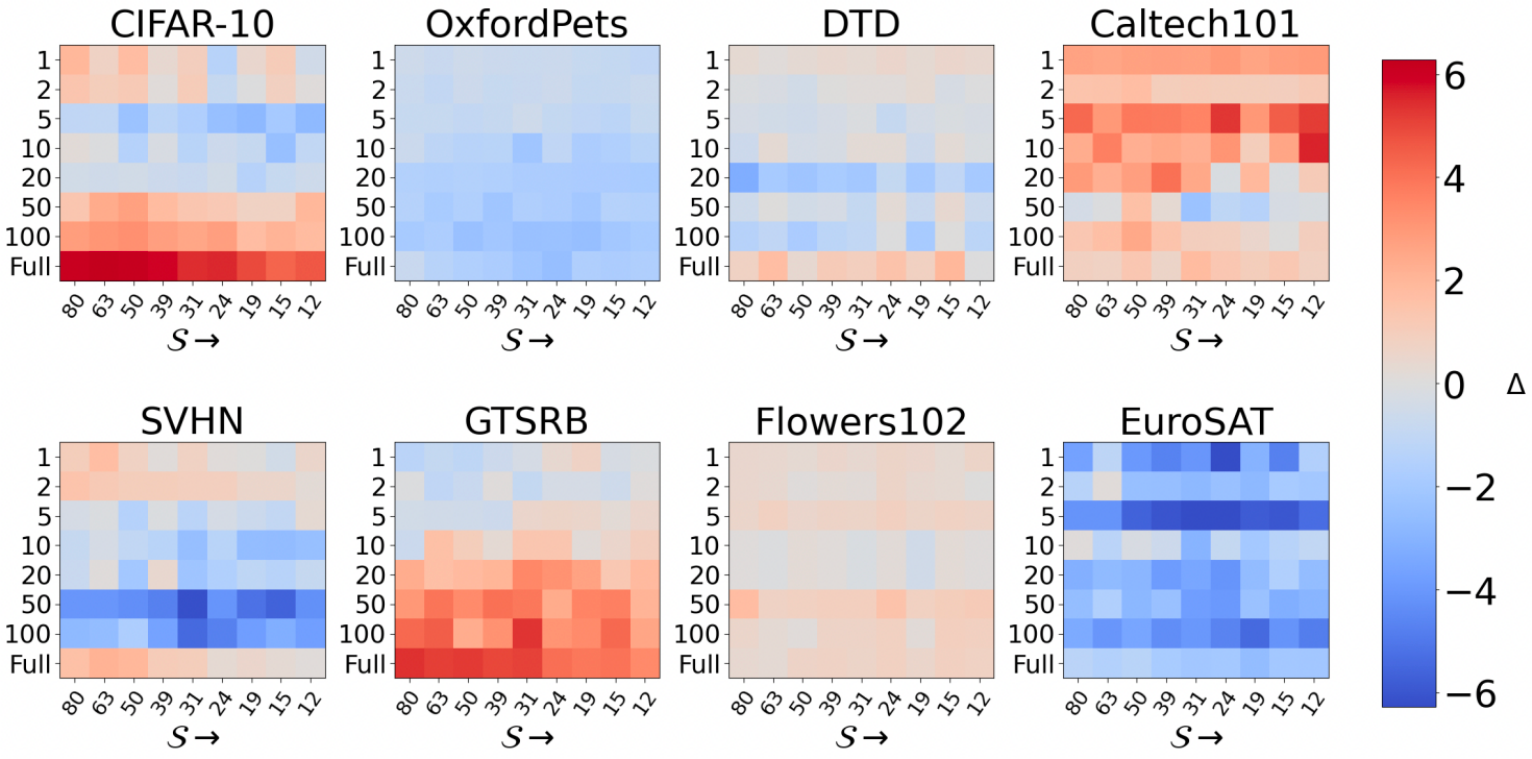}
    \caption{\textbf{Performance of LT solutions on Transfer via RLM-VP.} $\Delta-$difference for RLM-VP transfer of LTH solutions. $\mathcal{S}$ represents increasing levels of \%-sparsity levels of Dense (leftmost), $79.122, ~62.575, ~49.35, ~38.793, ~30.407$, $23.807, ~18.664, ~14.671,$ and $11.571$ (rightmost).}
    \label{fig:rlm_appendix_results}
    \vspace{-2mm}
\end{figure}

Subsequently, we examine the transfer performance using RLM-VP in Figure \ref{fig:rlm_appendix_results} across the eight datasets. It is essential to highlight, based on the accuracy plots encompassing various model architectures, data volumes, and sparsity settings thus far, that RLM-VP consistently emerges as the least effective visual prompting (VP) method. It is frequently outperformed by a considerable margin compared to both FLM-VP and ILM-VP.

Across almost all datasets, with the exceptions of EuroSAT and GTSRB, the performance trends indicate a close similarity between the lottery ticket hypothesis (LTH) solutions and dense models, with sparse model transfer generally resulting in performance deterioration. In EuroSAT, similar to transfers using ILM-VP and FLM-VP, even under RLM-VP, the sparse model outperforms the dense model by approximately 2-5\%.

In summary, considering the LTH solution trends presented in both the main text and this section, it becomes evident that, for the state-of-the-art visual prompting (VP) method ILM-VP and, for the most part, across various data and sparsity configurations of FLM-VP, sparse model transfer typically leads to performance degradation compared to their dense counterparts. For RLM-VP, a mixed trend is observed, though the interpretation is challenging due to the consistently poor transfer performance under this VP method.

\section{Additional CLIP Experiments}\label{sec:clip_appendix}

\begin{figure}[!h]
    \centering   \includegraphics[width=0.99\linewidth]{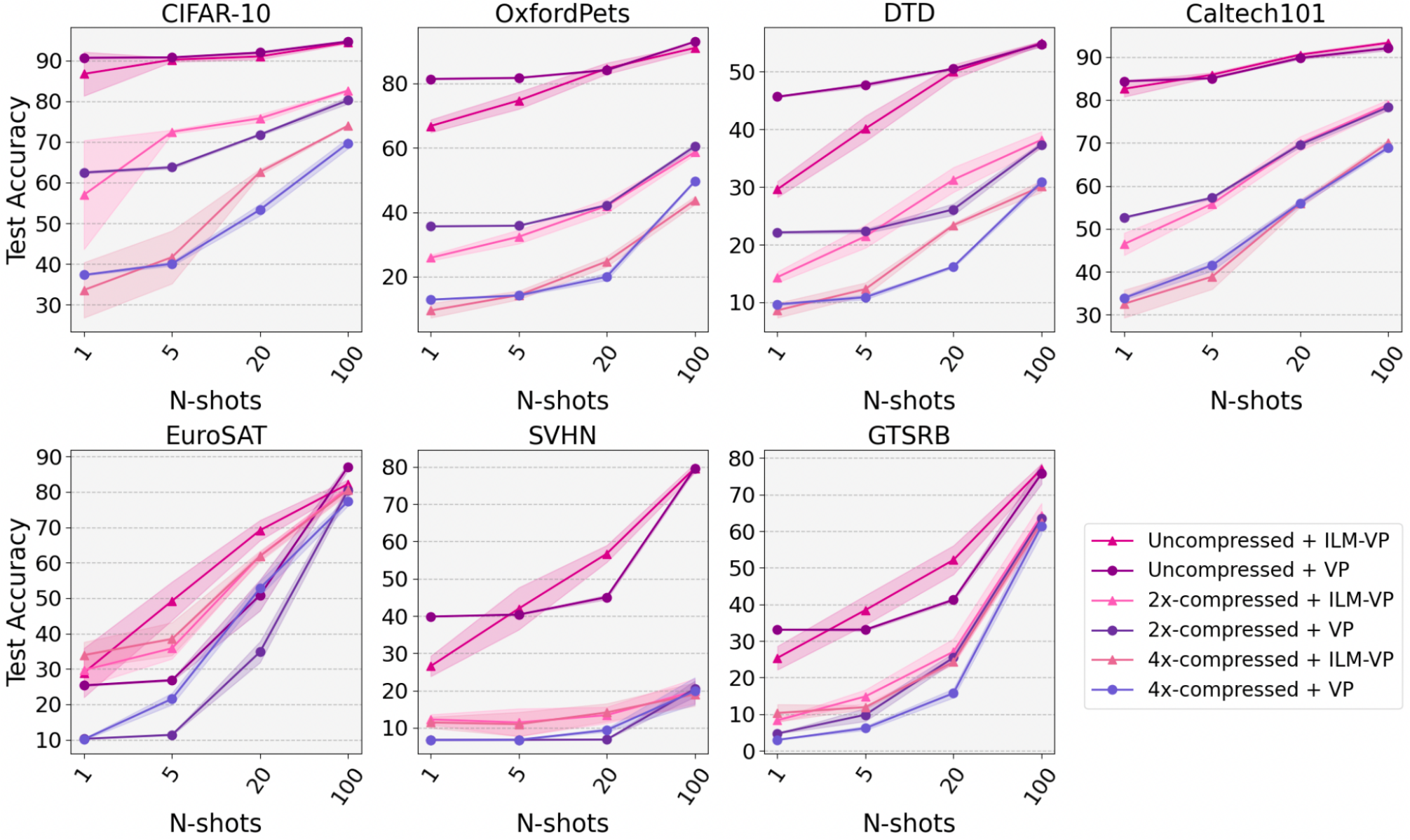}
    \caption{Transfer performance of uncompressed and compressed variants of CLIP across different N-shots configurations on a range of downstream datasets.}
    \label{fig:clip_low_data}
    \vspace{-2mm}
\end{figure}

In this section, we extend our analysis beyond the results presented in \cref{sec:sparse_foundation_models} of the main text, where the reported results pertain only to the full data volume setting. For consistency with the settings of the other experiments detailed in this manuscript, we apply the same N-shot variability set-up to three different variants of CLIP \cite{radford2021learning} - uncompressed, 2x compressed and 4x compressed - obtained through the UPop \cite{shi2023upop} compression method. Using the identical hyperparameter configuration outlined in Section \cref{sec:sparse_foundation_models}, the results presented in Figure \ref{fig:clip_low_data} reaffirm the trends observed in the full data volume setting from the main text. Across all N-shot settings within the seven downstream datasets used, a consistent decline in performance is observed when transitioning from the uncompressed CLIP variant to the 2x and 4x compressed variants.

With the exception of EuroSAT, there is a statistically significant performance gap between the uncompressed and compressed variants across all datasets. Across all settings, the performance follows the pattern: uncompressed variant $\gg$ 2x compressed variant $>$ 4x compressed variant. For example, in the OxfordPets dataset, we note a decrease in accuracy exceeding 40\% and 65\% when comparing the uncompressed variant to the 2x compressed and 4x compressed variants, respectively, at N-shot values of 1 and 5.

\subsubsection{What Leads to the Hidden Cost in VP?}\label{sec:result_analysis}

The aforementioned results on different model compression methods and sparse vision models unveil the existence of a common weakness in the severely degraded performance of VP. We hypothesize that the observed degradation is a hidden cost of model compression that accidentally weakens the label-mapping capability of the original model in VP. To verify this hypothesis, we conduct the following experiments and analyses to track the changes of VP in label mapping and training dynamics. 

To explore the differences between visual prompting with compressed models and their dense, full-precision counterparts, extending beyond accuracy, we begin by examining the label mapping process under ILM-VP for a ResNet-50 LT ($\mathcal{S}=14.671$) in a few-shot setting (N=5) on the \textup{OxfordPets} \cite{parkhi12a} and \textup{DTD} \cite{cimpoi14describing} datasets. As depicted in Figure \ref{fig:map}, our analysis reveals that the dense model maps the `Bombay' class from the \textup{OxfordPets} dataset to the `Schipperke' class from the \textup{ImageNet} \cite{deng2009imagenet} dataset, establishing a semantically closer mapping. On the contrary, the sparse LT model maps the same class to the `Carton' class from the \textup{ImageNet} dataset, a less semantically related mapping.

Furthermore, for the target dataset \textup{DTD}, the dense model maps the `Zig-Zagged' class to the `Chiffonier' class of the dataset \textup{ImageNet}. While the object categories do not directly correspond, it can be argued that zig-zag textures are more prevalent on the furniture texture frames of chiffoniers, as evident from the second example of the `Zig-Zagged' class compared to the third example of the `Chiffonier' class. This highlights a critical drawback of sparse models, indicating that they suffer from inferior label mapping, which ultimately hinders downstream performance.

A comparable pattern emerges in the context of model quantization. For example, in the case of the target class "Sphynx" (a cat breed) from the \textup{OxfordPets} dataset, the 2-bit quantized version of DeiT incorrectly assigns it to the unrelated label "tub," whereas the full-precision 32-bit variant of DeiT accurately maps it to the semantically related class "Mexican Hairless" (a dog breed) from the source \textup{ImageNet} dataset.


\begin{figure}[!h]
    \centering   \includegraphics[width=0.99\linewidth]{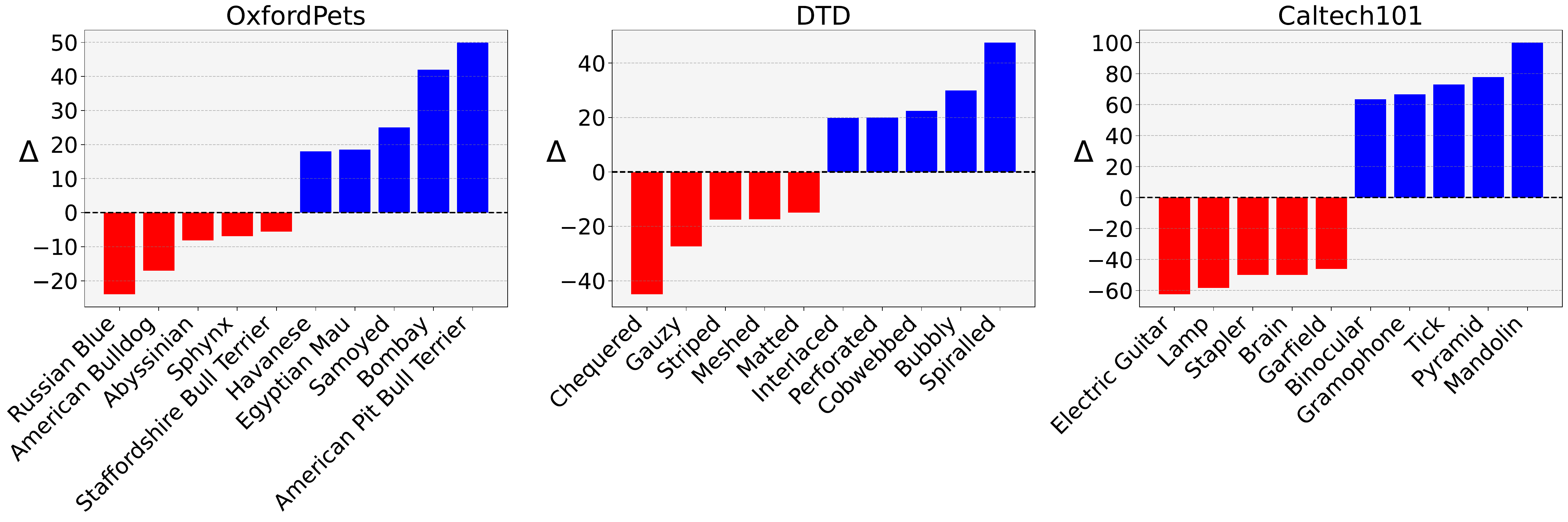}
    \caption{\textbf{Class-wise Performance Analysis.} Classes with Top-5 drop and gains in accuracy of transfer via ILM-VP on the three datasets of OxfordPets, DTD, and Caltech101. $\Delta$ represents the mean difference in top-1 accuracy of the dense model and the sparse model for each of the classes.}
    \label{fig:class_wise_analysis}
    \vspace{-2mm}
\end{figure} 

\begin{figure}[t]
    \centering   
    \begin{subfigure}{\linewidth}
    \includegraphics[width=\linewidth]{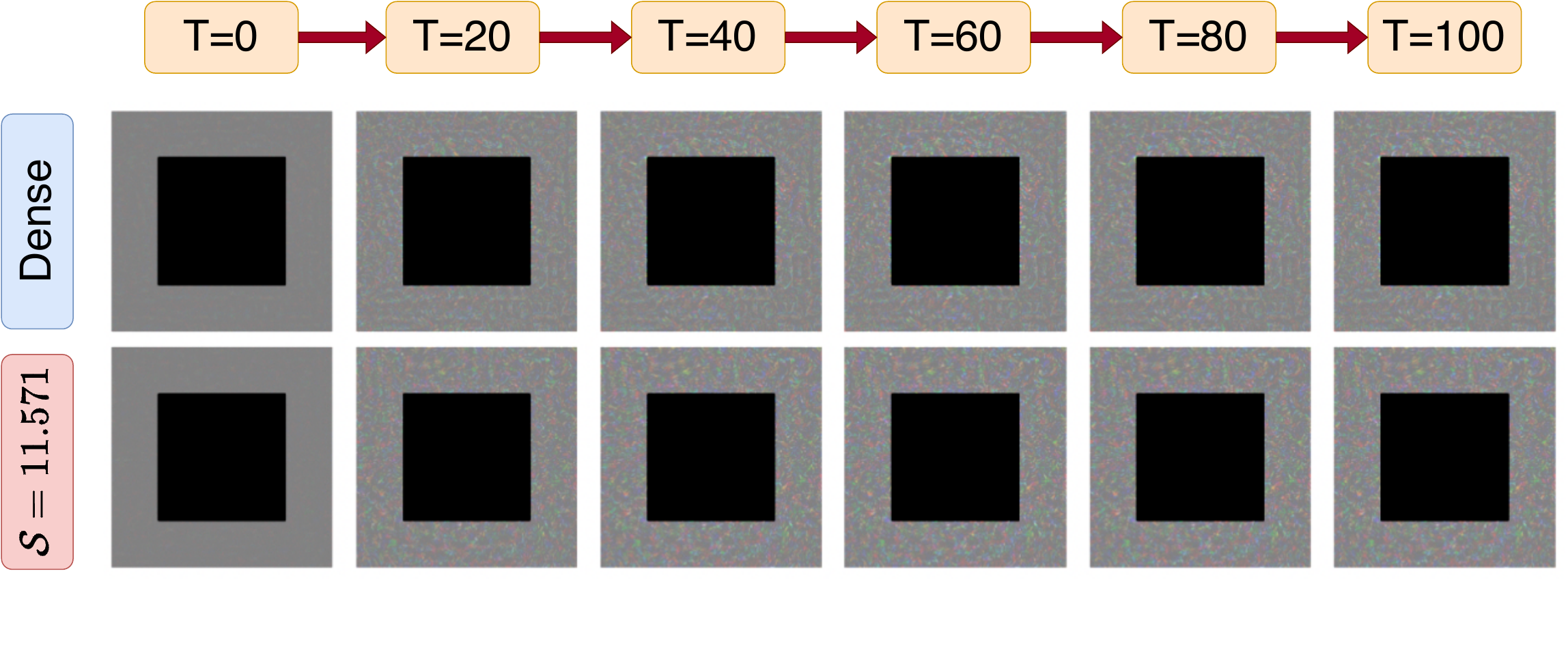}
    \caption{\textbf{Evolution of VP.} Visual prompt pattern versus number of training epoch for a ResNet-50 dense model and sparse LTH solution ($\mathcal{S}=11.571\%$). (Best viewed when zoomed in)}
    \end{subfigure}
    
    \bigskip
    
    \begin{subfigure}{\linewidth}
    \includegraphics[width=\linewidth]{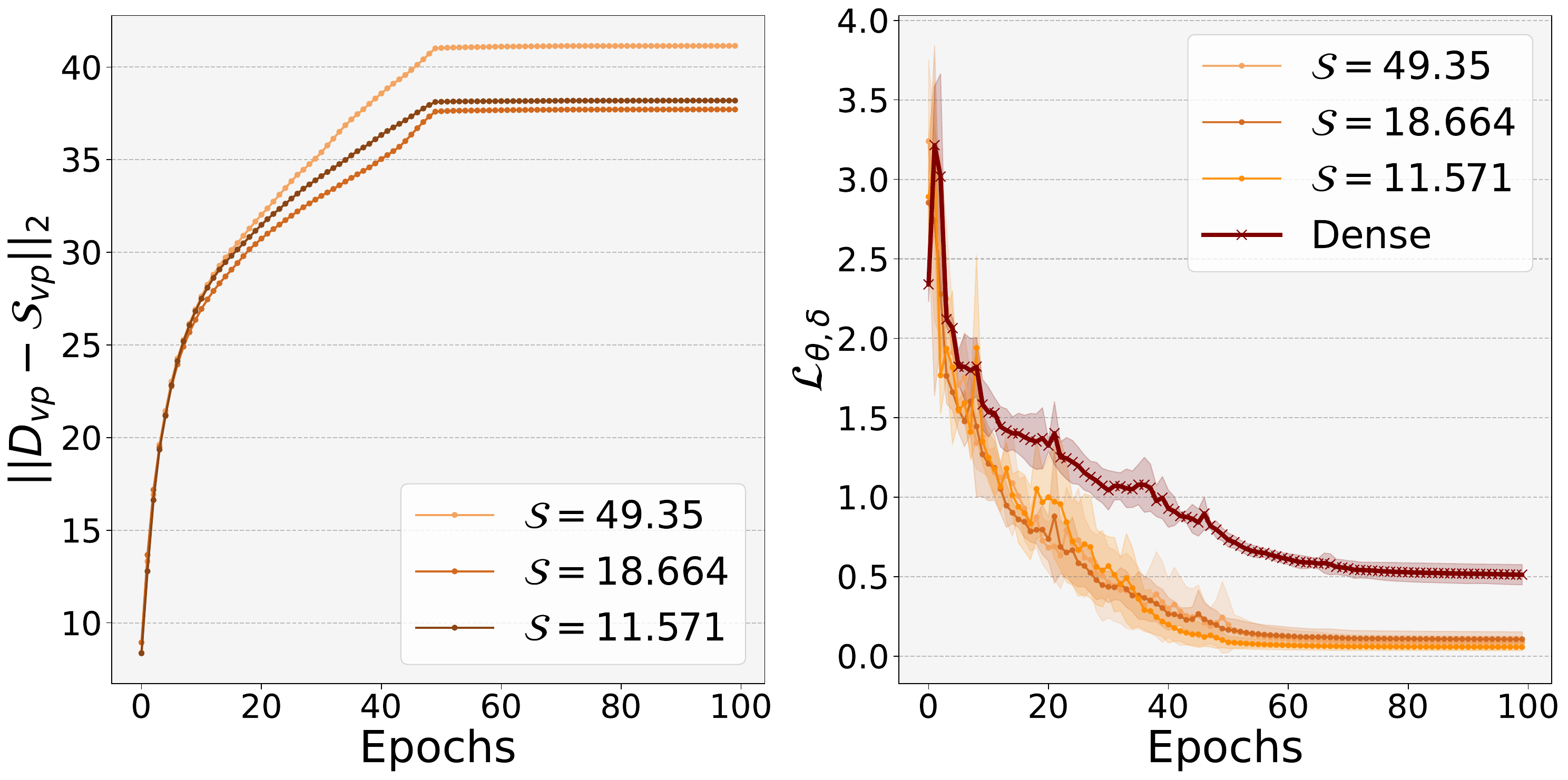}
    \caption{From left to right: (i) Change in $L_2$-norm of the difference between the learned visual prompts of a dense ResNet-50 model and its sparse LTH solutions variants at varying sparsity. (ii) \textbf{Convergence Analysis}: Training loss trajectories of the dense and sparse LTH solutions.}
    \end{subfigure}
    \caption{Training dynamics of Visual Prompting for ResNet-50 dense and sparse LTH variants on the \textup{Caltech101} dataset under few-shot settings (N=5).}
    \label{fig:dynamics}
    \vspace{-2mm}
\end{figure} 

To accurately characterize and distinguish the class-wise performance of visual prompted sparse and dense models, we conduct an analysis across three datasets: (a) \textup{OxfordPets} \cite{parkhi12a}, \textup{DTD} \cite{cimpoi14describing}, and \textup{Caltech101} \cite{li_andreeto_ranzato_perona_2022}. Specifically, we identify the top five classes in which the sparse model outperforms the dense counterpart and vice versa. Using the sparse models featured in Figure \ref{fig:map}, we analyze visual prompted dense ResNet-50 and sparse LT ResNet-50 ($\mathcal{S}=14.671$) trained in a few-shot setting (N=5). As illustrated in Figure \ref{fig:class_wise_analysis}, our results indicate that while certain classes exhibit improved performance with the sparse model, the magnitude of this improvement is often less pronounced compared to the performance advantage of the dense model. For example, in the case of the \textup{Caltech101} dataset, the dense model achieves a 100\% increase in accuracy for the `Mandolin' class when compared to the sparse LT model. Although these findings offer preliminary insights into the class-wise performance dynamics between sparse and dense models, a more comprehensive understanding of these dynamics is deferred to future investigations.

Finally, we examine the evolutionary trajectories and training dynamics of the visual prompts learned by sparse models compared to their dense counterparts. To conduct this analysis, we consider both ResNet-50 dense and sparse LTH solutions ($\mathcal{S} = 11.571, 18.664, 49.35$) trained on the \textup{Caltech101} dataset in a few-shot setting (N=5). The insights gained from this examination are illustrated in Figure \ref{fig:dynamics}.

As training progresses, we observe a gradual increase in the $L_2$-norm of the difference between the visual prompt learned by the dense model and each of the sparse model variants. This analysis demonstrates how the visual prompt of the sparse models with more compute diverges from the more optimal visual prompt learned by the dense model.

\section{Additional Experiments }

\begin{figure}[!h]
    \centering   \includegraphics[width=0.99\linewidth]{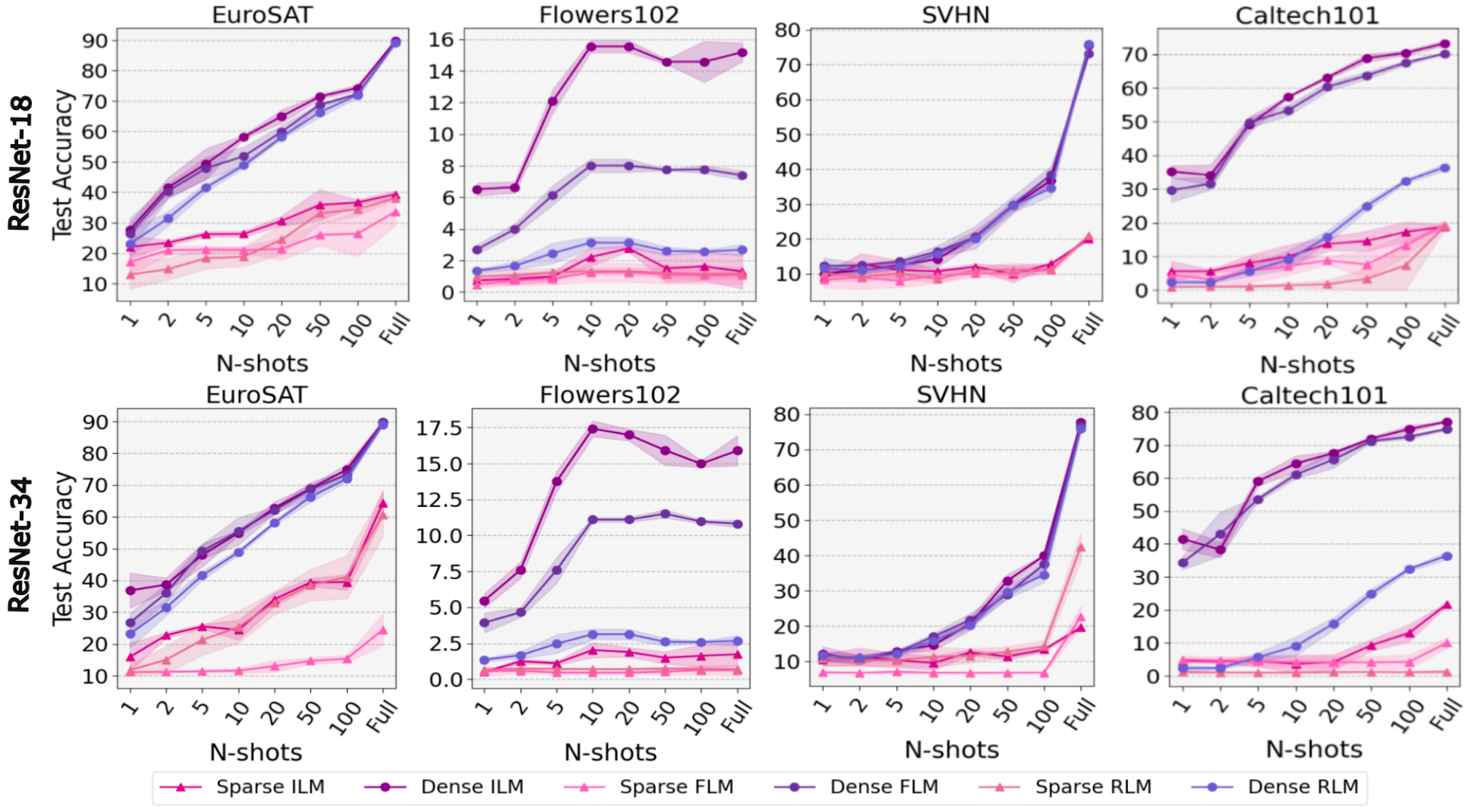}
    \caption{\textbf{GMP-pruned ResNet-18/34.} Transfer performance measured by test accuracy of pruned ResNet-18/34 model on a variety of downstream datasets and varying levels of data budgets.}
    \label{fig:resnet_gmp_supp}
    \vspace{-2mm}
\end{figure}

\begin{figure}[!h]
    \centering   \includegraphics[width=0.99\linewidth]{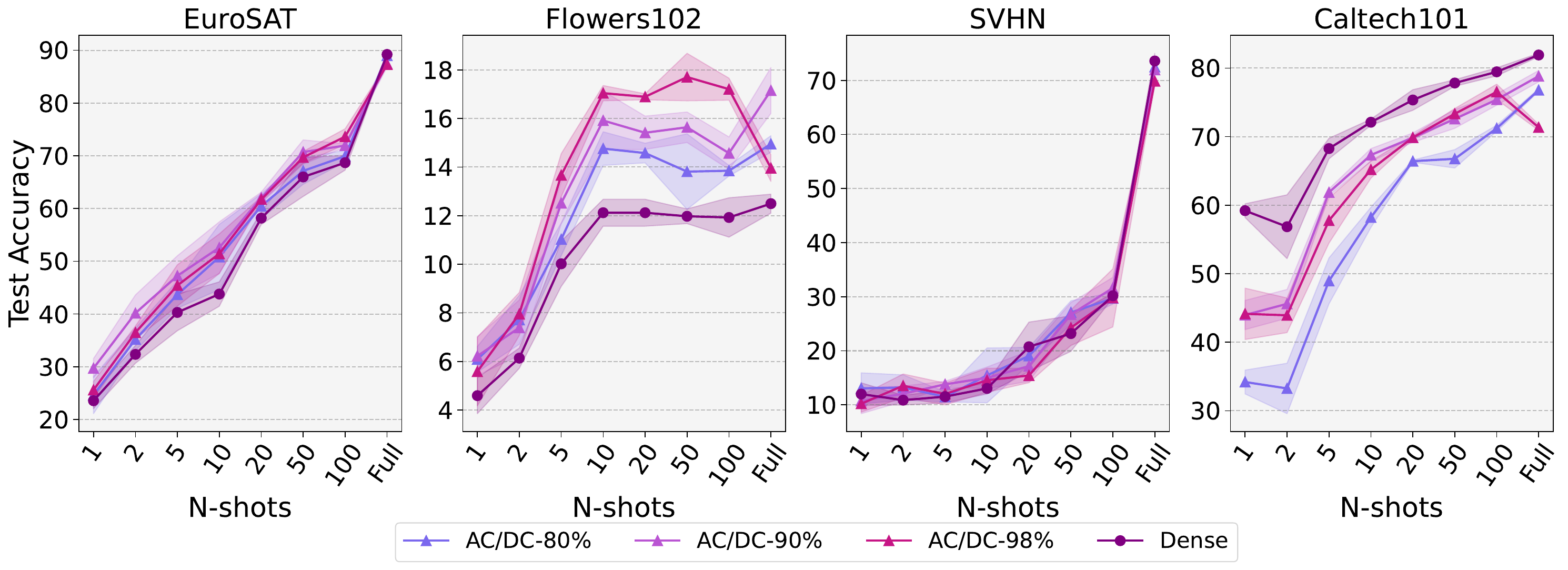}
    \caption{\textbf{AC/DC-pruned ResNet-50.} Transfer performance measured by test accuracy of pruned ResNet-50 model on a variety of downstream datasets and varying levels of data budgets.}
    \label{fig:acdc_supp}
    \vspace{-2mm}
\end{figure}

\begin{figure}[!h]
    \centering   \includegraphics[width=0.99\linewidth]{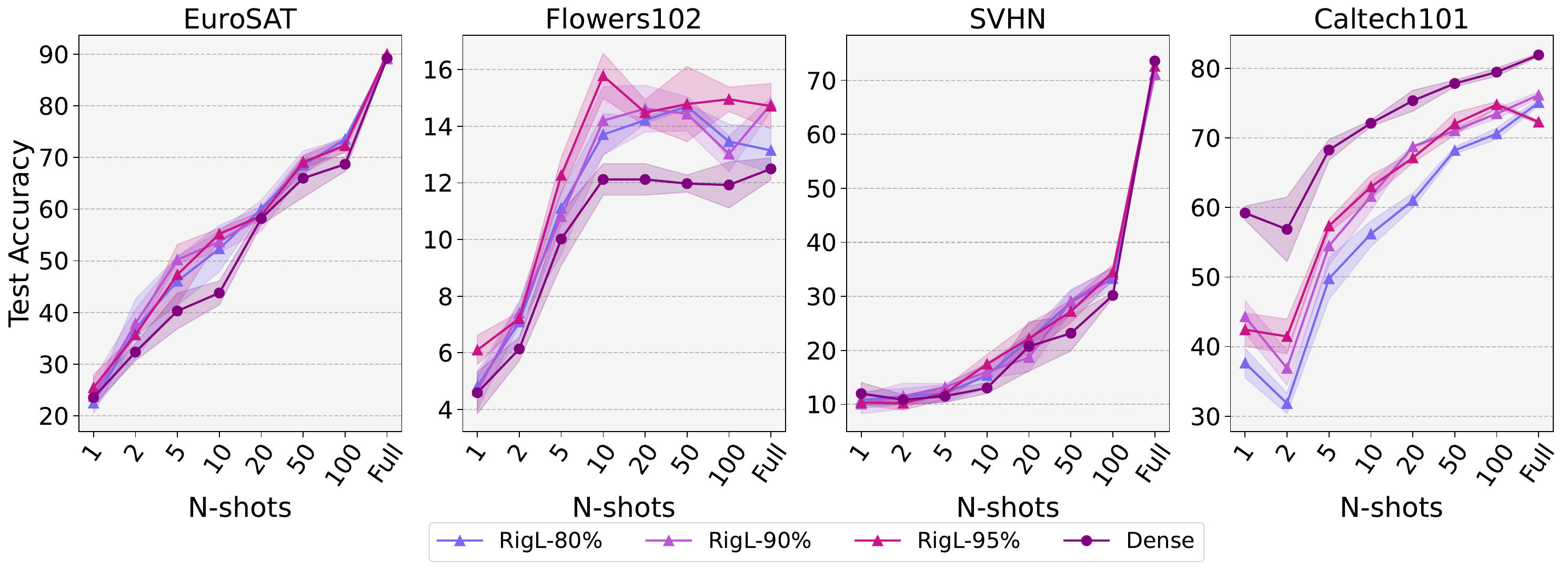}
    \caption{\textbf{RigL-pruned ResNet-50.} Transfer performance measured by test accuracy of pruned ResNet-50 model on a variety of downstream datasets and varying levels of data budgets.}
    \label{fig:rigl_supp}
    \vspace{-2mm}
\end{figure} 

\begin{figure}[h!]
\centering
    \small
\includegraphics[width=0.99\linewidth]{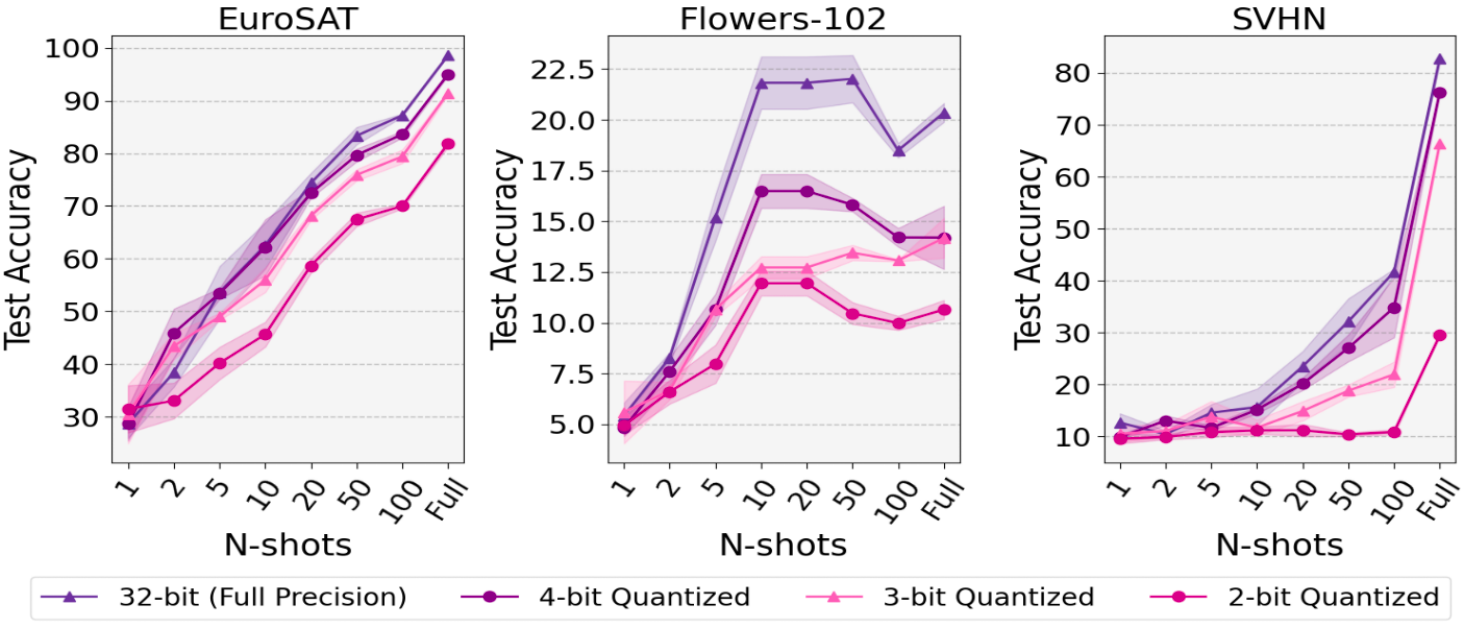}
    \caption{\textbf{VVTQuantized DeiT-T.} Transfer performance measured by test accuracy of quantized DeiT-T models on a variety of downstream datasets and varying levels of data budgets.}
    \vspace{-2mm}
    \label{fig:vvtq_deit_supp}
\end{figure}

\begin{figure}[h!]
\centering
    \small
\includegraphics[width=0.99\linewidth]{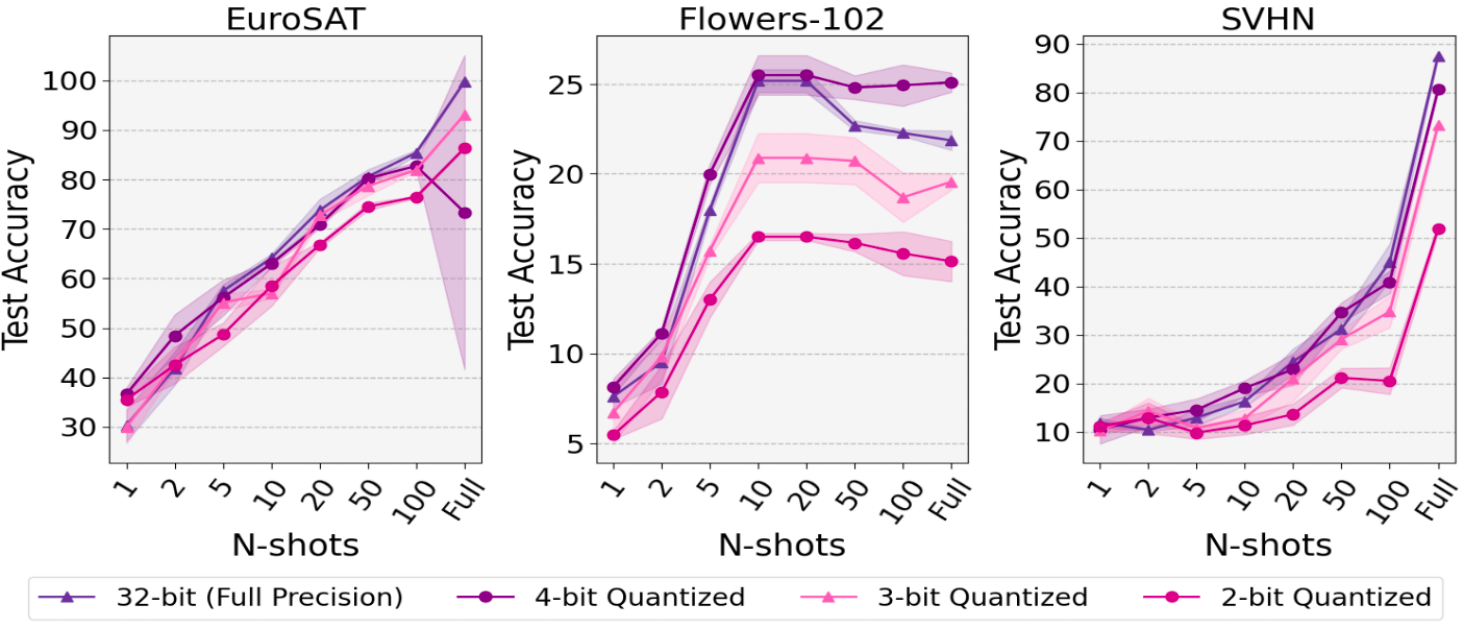}
    \caption{\textbf{VVTQuantized Swin-T.} Transfer performance measured by test accuracy of quantized Swin-T models on a variety of downstream datasets and varying levels of data budgets.}
    \vspace{-2mm}
    \label{fig:vvtq_swin_supp}
\end{figure}

\end{document}